\documentclass{article}

\usepackage{arxiv}

\usepackage[utf8]{inputenc} % allow utf-8 input
\usepackage[T1]{fontenc}    % use 8-bit T1 fonts
\usepackage{hyperref}       % hyperlinks
\usepackage{url}            % simple URL typesetting
\usepackage{booktabs}       % professional-quality tables
\usepackage{amsfonts}       % blackboard math symbols
\usepackage{nicefrac}       % compact symbols for 1/2, etc.
\usepackage{microtype}      % microtypography
\usepackage{lipsum}
\usepackage{graphicx}
\usepackage{subcaption}
\usepackage{graphicx}
\usepackage[export]{adjustbox}
\usepackage{color}
\graphicspath{ {./images/} }

\usepackage{authblk} % 用于处理多作者和隶属关系
\usepackage{hyperref} % 用于创建超链接

\title{ContourFormer:Real-Time Contour-Based End-to-End Instance Segmentation Transformer}



\author{Weiwei Yao}
\author{Chen Li}
\author{Minjun Xiong}
\author{Wenbo Dong}
\author{Hao Chen}
\author{Xiong xiao}

\affil{Zhuzhou CRRC Times Electric Co., Ltd., China}

\begin{document}
\maketitle

\renewcommand{\thefootnote}{\fnsymbol{footnote}}
\footnotetext[1]{Email:yaoww1@csrzic.com.}
\footnotetext[2]{The code is available at: \url{https://github.com/talebolano/Contourformer}}

\begin{abstract}
This paper presents Contourformer, a real-time contour-based instance segmentation algorithm. The method is fully based on the DETR paradigm and achieves end-to-end inference through iterative and progressive mechanisms to optimize contours. To improve efficiency and accuracy, we develop two novel techniques: sub-contour decoupling mechanisms and contour fine-grained distribution refinement. In the sub-contour decoupling mechanism, we propose a deformable attention-based module that adaptively selects sampling regions based on the current predicted contour, enabling more effective capturing of object boundary information. Additionally, we design a multi-stage optimization process to enhance segmentation precision by progressively refining sub-contours. The contour fine-grained distribution refinement technique aims to further improve the ability to express fine details of contours. These innovations enable Contourformer to achieve stable and precise segmentation for each instance while maintaining real-time performance. Extensive experiments demonstrate the superior performance of Contourformer on multiple benchmark datasets, including SBD, COCO, and KINS. We conduct comprehensive evaluations and comparisons with existing state-of-the-art methods, showing significant improvements in both accuracy and inference speed. This work provides a new solution for contour-based instance segmentation tasks and lays a foundation for future research, with the potential to become a strong baseline method in this field. 
\end{abstract}

% keywords can be removed

\section{Introduction}
Instance segmentation \cite{he2017mask} is a fundamental task in the field of computer vision, aiming to precisely identify each individual object and its corresponding contour within an image. In recent years, instance segmentation technology has received widespread attention and achieved significant breakthroughs in research. The studies in this domain have primarily focused on two paradigms: mask-based methods and contour-based methods. The former identifies target regions through pixel-level masks, while the latter delineates object contours using sparse boundary points.

Currently, mask-based methods are the mainstream choice and have achieved significant improvements in prediction accuracy. However, these methods require precise pixel-level predictions for each object, which often results in high computational resource consumption and memory usage, making them difficult to apply in real-time scenarios. Additionally, mask-based methods lack explicit structure modeling, leading to potential uncertainties or noise in boundary detail delineation and internal region processing. On the other hand, contour-based methods have alleviated computational resource occupancy to some extent but still fall short of achieving parity with mask-based methods, particularly in large-scale benchmark tests. Researchers in the deep learning community have proposed various improvement strategies. However, these methods\cite{8953834,zhang2022e2ec,xie2020polarmask,peng2020deep,liu2021dance,feng2024recurrent} still struggle to match the performance of state-of-the-art mask-based methods, especially when dealing with complex scenarios.

To address these challenges, we propose a novel instance segmentation method called Contourformer. This approach is based on the DETR \cite{carion2020end} paradigm, inheriting its advantages such as obtaining final results in an end-to-end manner without relying on Non-Maximum Suppression (NMS) or anchor boxes. Building upon DETR, Contourformer directly predicts object boundaries as polygons and employs iterative and progressive mechanisms to refine initial contours by estimating residual displacements, thereby delineating the target instances accurately.Our model demonstrates state-of-the-art performance on the SBD \cite{hariharan2011semantic} dataset, MS-COCO \cite{lin2014microsoft} dataset, and KINS \cite{qi2019amodal} dataset. For images of size 512×512, Contourformer achieves an inference speed of 24.6 frames per second (fps) on an NVIDIA A30 GPU, with accuracy significantly surpassing existing contour-based instance segmentation algorithms.The primary contributions of our work are as follows:
\begin{itemize}
\item DETR paradigm-based contour regression model: Our method is fully grounded in the DETR paradigm, employing iterative and progressive mechanisms to optimize contours, thereby enabling end-to-end inference.
\item Innovative sub-contour decoupling mechanism: We propose an innovative sub-contour decoupling mechanism that decomposes complex overall contours into multiple local sub-parts. By independently modeling the geometric features of each sub-part, this approach reduces the difficulty of learning while ensuring model stability across various scales and viewpoints.
\item Redefining contour regression within the decoder stage: We redefine contour regression during the decoder stage, allowing the model to progressively refine location estimates, which significantly enhances the final localization accuracy.
\end{itemize}

\section{Related Work}
\label{sec:headings}
This paper reviews the research progress in the field of instance segmentation from the perspectives of mask-based and contour-based approaches, and provides a detailed analysis of various representative methods.
\paragraph{Mask-based instance segmentation.}
In the domain of mask-based instance segmentation, mainstream research can be broadly categorized into two main paradigms: methods that rely on bounding boxes and those that directly predict masks. Early approaches, such as Mask R-CNN \cite{he2017mask}, adhered to a two-stage paradigm where bounding box detection was followed by mask prediction within each detected region. Subsequent advancements introduced cascaded integration techniques, exemplified by HTC \cite{chen2019hybrid} and RefineMask \cite{Zhang_2021_CVPR}, which interleaved detection and segmentation features or fused instance features across different stages to refine mask predictions.To address the complexity of mask representation, DCT-Mask \cite{shen2021dct} proposed a DCT-based mask representation, while Patch-DCT \cite{wen2022patchdct} further enhanced precision by dividing masks into independent patches and applying DCT-based representation at the patch level. 
More recently, methods have shifted away from bounding box reliance, instead focusing on direct mask prediction. YOLACT \cite{yolact-iccv2019} achieved instance segmentation by generating prototype masks and their combination coefficients. BlendMask\cite{9156939} predicted 2D attention maps for each proposal and combined them with ROI features to accomplish the segmentation task. SOLO \cite{wang2020solo} introduced dynamic convolutional kernels to construct high-resolution instance segmentation methods. K-Net \cite{zhang2021k} integrated Transformer structures, gradually optimizing masks through a progressive approach. MaskFormer \cite{cheng2021per} combined the DETR paradigm with dynamic convolutional kernels to separately predict class labels and mask results. Mask2former \cite{cheng2022masked} adopted Deformable-detr \cite{zhudeformable} as the base model, restricting cross-attention to predicted mask regions for local feature extraction. MaskDino \cite{li2023mask} merged DINO\cite{zhang2022dino}, an object detection algorithm that incorporates various training acceleration methods\cite{li2022dn,liu2022dabdetr}, with dynamic convolutional kernels for generating masks, enabling simultaneous output of both masks and bounding boxes.
\paragraph{Contour-based instance segmentation.}
In contrast to mask-based approaches, contour-based methods focus on predicting object boundaries or polygonal representations of instances. These methods can be broadly categorized into two types: those using polar coordinates and those employing Cartesian coordinates. For instance, PolarMask \cite{xie2020polarmask} reformulated instance segmentation as a contour regression problem in polar coordinates, while DeepSnake\cite{peng2020deep} initializes contours using bounding box predictions, followed by multiple deformation steps to complete the segmentation task. Building on DeepSnake, Dance \cite{liu2021dance} improves the matching scheme between predicted and target contours and introduces an attention deformation mechanism. E2EC proposes a learnable contour initialization architecture, significantly enhancing performance. Polysnake \cite{feng2024recurrent} presents a more lightweight deformable contour module along with shape loss to encourage and regulate object shape learning. BoundaryFormer \cite{lazarow2022instance} introduces a differentiable rasterization module, using pixel-level masks to supervise model training. PolyFormer \cite{liu2023polyformer} adopts a Seq2Seq\cite{sutskever2014sequence} framework, taking image patches and text query tokens as input and autoregressively outputting sequences of polygon vertices.

Different from existing methods, Contourformer fully adopts the DETR paradigm and employs iterative and progressive mechanisms to learn object contours. It transforms polygon regression from predicting fixed coordinates to modeling probability distributions, thereby achieving precise and robust estimation for various objects.

\section{Methodology}

\begin{figure}
  \centering
  \includegraphics[width=1.0\textwidth]{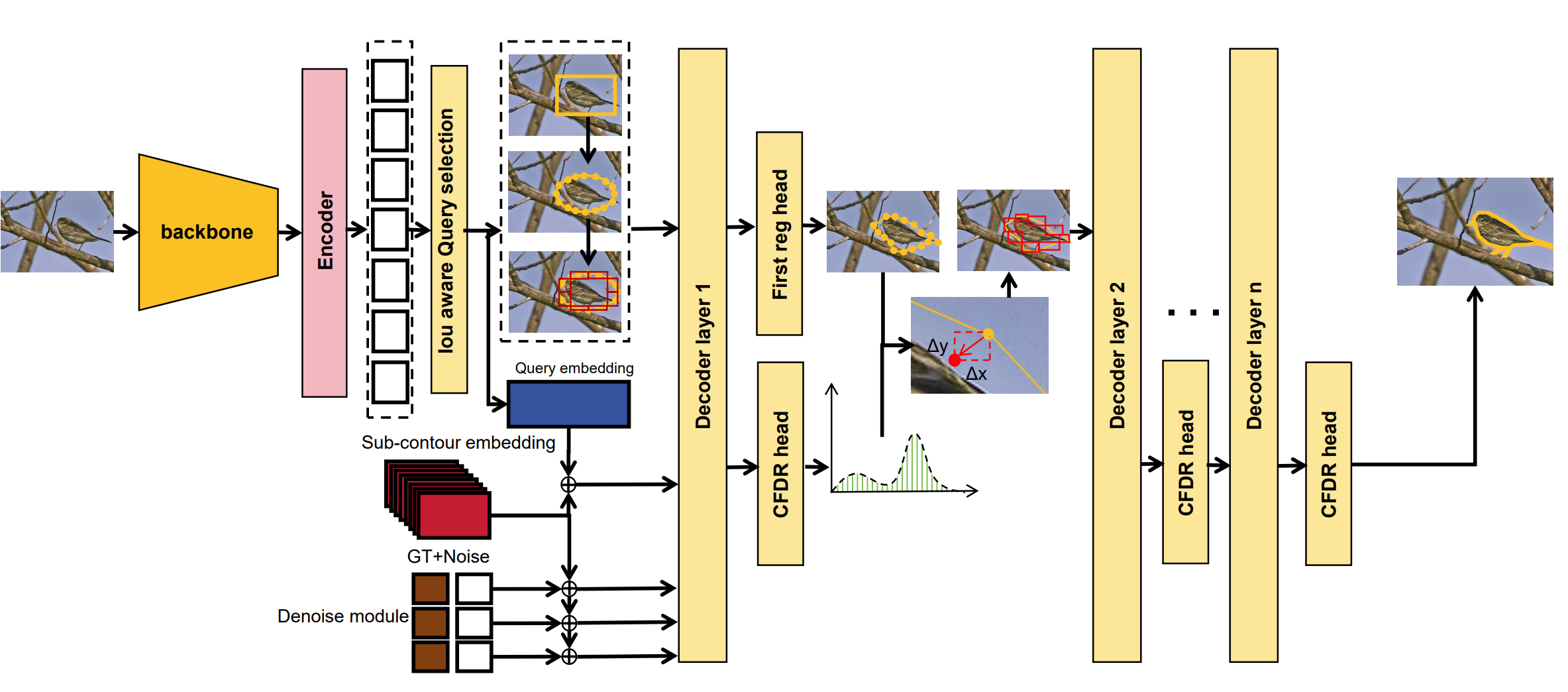}
  \caption{ Overview of Contourformer: Given an image, multi-scale features are collected through the Backbone and Encoder. Initial bounding boxes and Query features for each object are proposed. A simple ellipse is used to initialize a polygon contour for each target, which is then divided into eight sub-contours. Corresponding bounding boxes are created for these sub-contours. N stacked Transformer decoder layers iteratively refine each sub-contour, with the bounding box of each sub-contour providing the feature extraction range for cross-attention. The FDR head provides fine-grained distributions for each boundary point during iterations. The network employs a denoising module to accelerate training and enhance accuracy.}
  \label{fig:fig1}
\end{figure}

\label{sec:headings}
The proposed Contourformer framework is shown in figure \ref{fig:fig1}. This framework is built upon the D-FINE \cite{peng2024d} object detection model and extends the regression of bounding boxes to the regression of contours. To achieve efficient training, Contourformer employs an iterative method for contour deformation and introduces a denoising mechanism to accelerate the convergence process. In implementation, we have redesigned the DETR decoder's iterative architecture to enable progressive refinement of contour estimates in each iteration. Through this iterative optimization, the contours can stably enclose the target objects, ultimately reaching a consistent and precise state.For better understanding and analysis, we divide the framework into two core modules: (1) Sub-contour Decoupling Mechanism: This module aims to effectively decouple complex contour relationships and enhance the model's ability to capture target boundaries; (2) Contour Fine-grained Distribution Refinement: This module further improves the precision of segmentation results through fine-grained probability modeling and distribution optimization.These two modules will be detailed in subsequent sections to illustrate their design concepts and implementation methods.

\subsection{Sub-contour decoupling mechanism}

\begin{figure}
  \centering
  \includegraphics[width=0.8\textwidth]{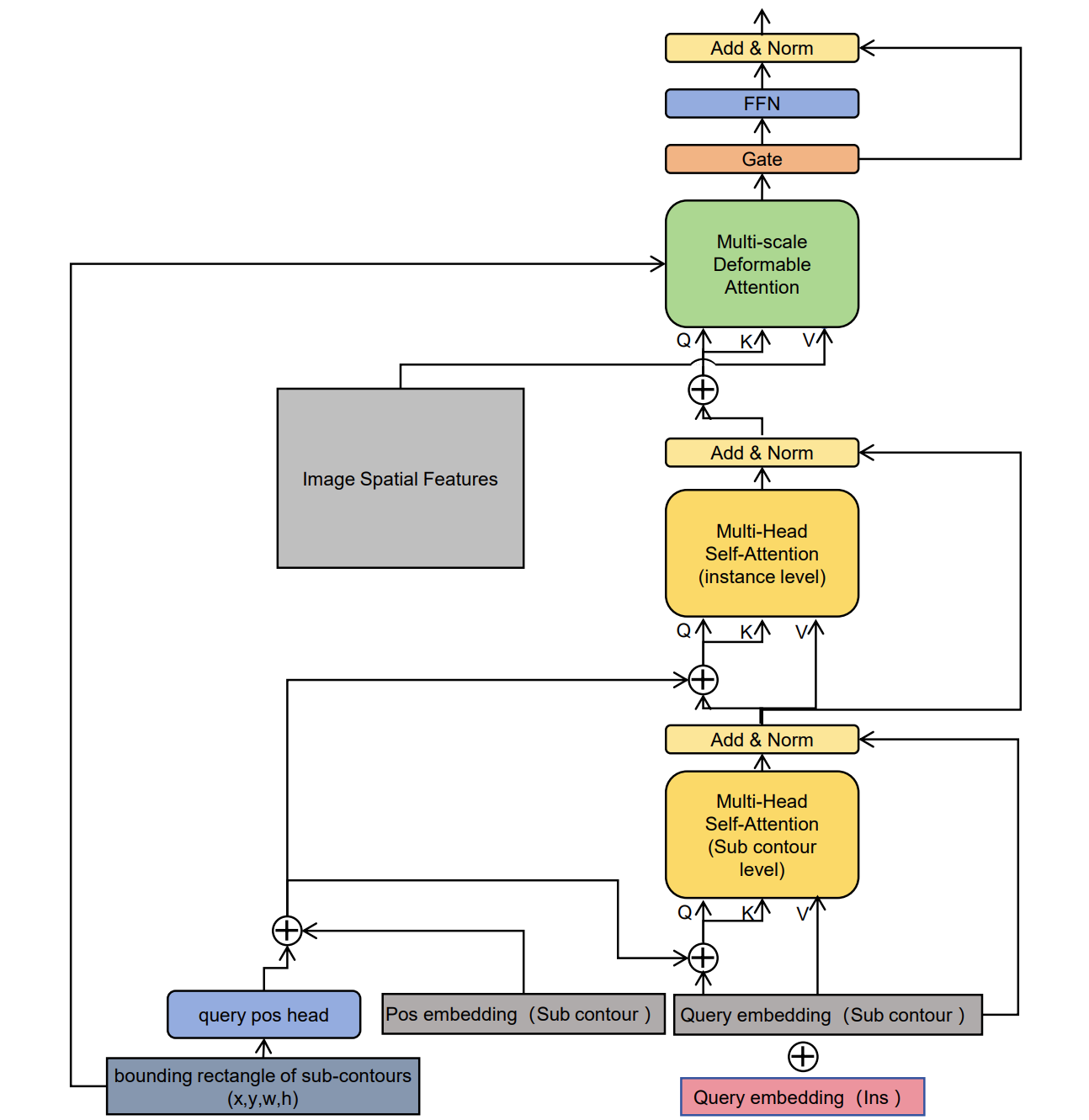}
  \caption{Sub-contour decoupling mechanism: Each decoder layer uses self-attention and cross-attention to update the queries. It performs two rounds of self-attention, first among $\left\{q_{j}^{\mathrm{c}}\right\}_{j=0}^{N_{c}-1}$ and then among $\left\{q_{i}^{\mathrm{ins}}\right\}_{i=0}^{N_{q}-1}$. During cross-attention, the bounding box $\left\{x, y, w, h\right\}$ corresponding to the sub-contour $\left\{v_{i}\right\}_{i=0}^{N_{s}-1}$ predicted in the previous layer is used as the range for feature sampling and query updating.}
  \label{fig:fig2}
\end{figure}

The initial contour is given by Contourformer's encoder, referencing previous methods, and is determined by sampling from the inscribed ellipse of the initially predicted bounding box\cite{8953834}. The initial contour $V=\left\{v_{i}\right\}_{i=0}^{N_{v}-1}$, where $N_{v}$ is the number of boundary points for each contour, and each boundary point $v_{i}$ is represented by $(x_{i}, y_{i})$ , which means that the model ultimately needs to predict the values of $2N_{v}$. We divide the contour $\left\{v_{i}\right\}_{i=0}^{N_{v}-1}$ into $N_{c}$ sub-contour regions $\left\{v_{i}\right\}_{i=0}^{N_{s}-1}$ (where $N_{s}$ is the number of boundary points in each sub-contour, which can be divided by $N_{v}$,i.e., $N_{c} \times N_{s}=N_{v}$), with different sub-contour regions responsible for generating different parts of the contour. Queries for different sub-contour regions consist of two parts: in addition to a set of instance-level queries $\left\{q_{i}^{\mathrm{ins}}\right\}_{i=0}^{N_{q}-1}$ (where $N_{q}$ is the number of instances), representing each instance, and an additional set of sub-contour-level queries $\left\{q_{j}^{\mathrm{c}}\right\}_{j=0}^{N_{c}-1}$ (where $N_{c}$ is the number of sub-contours), shared among all instances. Each instance object corresponds to a set of sub-contour queries $\left\{q_{ij}^{\mathrm{sc}}\right\}_{j=0}^{N_{c}-1}$. The query formula for the $j$-th sub-contour of the $i$-th instance is:
\begin{equation}
q_{ij} = q_{i}^{\mathrm{ins}} + q_{j}^{\mathrm{c}}
\end{equation}
The decoder of Contourformer consists of multiple decoder layers, each of which uses self-attention and cross-attention to update the queries for sub-contour regions, as shown in Figure \ref{fig:fig2}. After dividing the contour regions, the number of queries involved in the computation in the decoder increases from $N_{q}$ to $N_{q}\times N_{c}$. Directly performing self-attention between $\left\{q_{ij}^{\mathrm{sc}}\right\}_{j=0}^{N_{c}-1}$ has a computational complexity of $O((N_{q}\times N_{c})^{2})$, which would significantly increase computational cost and memory consumption. Therefore, to achieve a balance between speed and performance, we adopt decoupled self-attention, i.e., perform two self-attentions: first, self-attention between $\left\{q_{j}^{\mathrm{c}}\right\}_{j=0}^{N_{c}-1}$, followed by attention between $\left\{q_{i}^{\mathrm{ins}}\right\}_{i=0}^{N_{q}-1}$.This reduces the computational complexity from $O((N_{q}\times N_{c})^{2})$ to $O((N_{q} + N_{c})^{2})$.During cross-attention, we use deformable attention mechanisms to make each query interact with input features. We use the outer bounding box $\left\{x, y, w, h\right\}$ corresponding to the predicted sub-contours $\left\{v_{i}\right\}_{i=0}^{N_{s}-1}$ from the previous layer to sample and update the queries. The restricted attention range can adapt to different sub-contour regions and capture contextual information useful for contour learning. The final predicted contours are formed by concatenating the individual sub-contours.

\subsection{Contour fine-grained distribution refinement}

\begin{figure}
  \centering
  \includegraphics[width=0.5\textwidth]{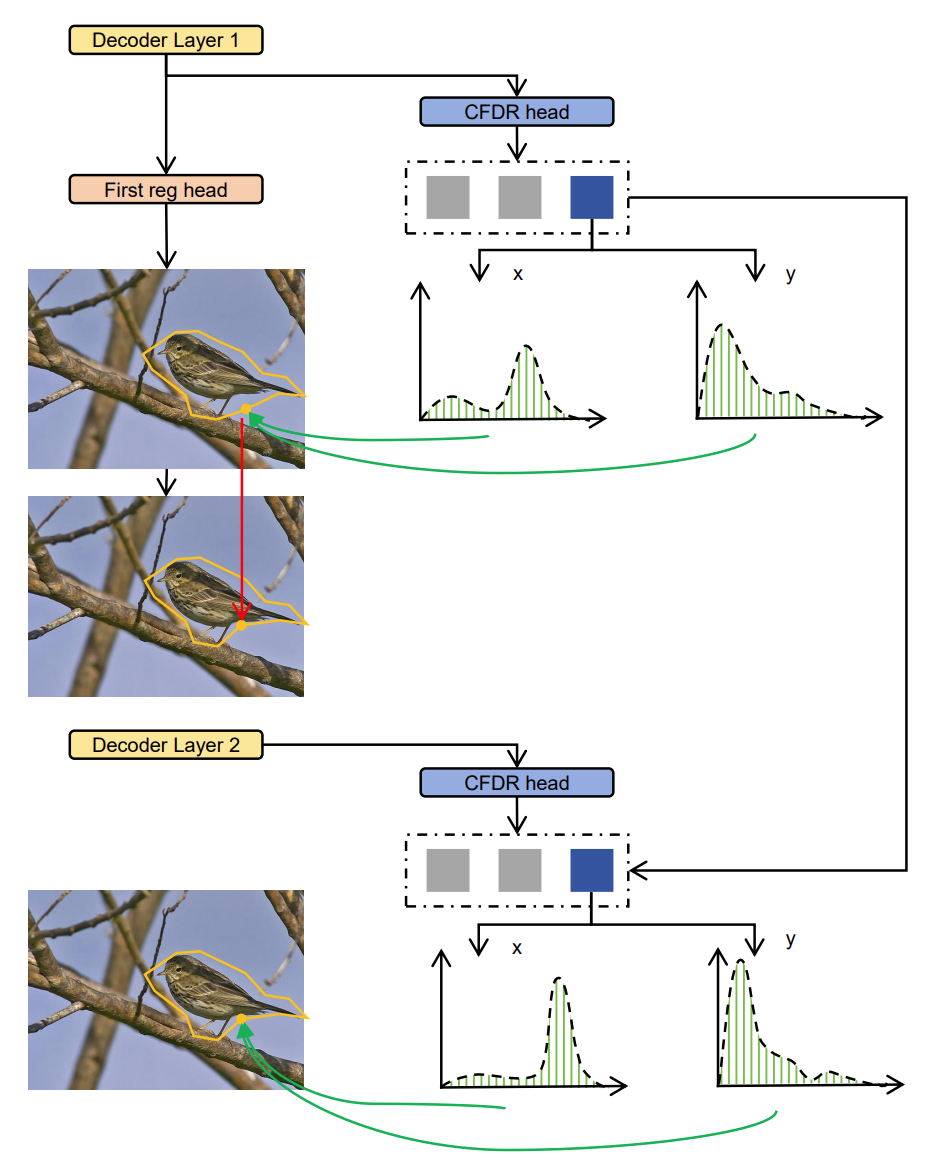}
  \caption{Contour Fine-Grained Distribution Refinement (CFDR): The first decoder layer predicts the initial contour and preliminary probability distribution using a conventional regression head and a CFDR head. Subsequently, each subsequent layer employs residual adjustments to update the probability distributions, resulting in more precise boundary localization.}
  \label{fig:fig3}
\end{figure}

Contour modeling in previous approaches has predominantly relied on deterministic representations, which inherently lack the capacity to capture positional uncertainty - particularly evident in ambiguous boundary scenarios. This rigid paradigm imposes two critical limitations: (1) constrained optimization flexibility due to fixed geometric constraints, and (2) amplified localization errors arising from minor prediction inaccuracies\cite{9792391}. Drawing inspiration from recent advancements in probabilistic bounding box estimation (D-FINE)\cite{peng2024d}, we propose Contour Fine-Grained Distribution Refinement (CFDR), a novel framework that extends fine-grained distribution modeling to contour-based object representation through multi-layer probabilistic refinement.

As illustrated in Figure \ref{fig:fig3}, the CFDR framework employs an iterative refinement process across successive decoder layers. The initial decoder layer establishes baseline predictions through two distinct components: (1)A conventional regression head producing the primary contour estimate $v^{0}$; (2)A probabilistic CFDR head generating preliminary offset distributions $\left\{Pr_{x}^{l}(n),Pr_{y}^{l}(n) \right\}$ for each boundary point.Subsequent layers perform progressive distribution refinement through residual adjustments, mathematically formulated as:
\begin{equation}
v^{l}=v^{0}+\left\{W_{c},H_{c}\right\}\cdot \sum_{n=0}^{N}W(n)Pr^{l}(n),l\in \left\{1,2,3...L \right\}
\end{equation}
Where $Pr^{l}(n)=\left\{Pr_{x}^{l}(n),Pr_{y}^{l}(n) \right\}$ represents the bivariate probability distributions over discrete offset bins for x/y coordinates. The refinement mechanism employs a residual learning paradigm:
\begin{equation}
Pr^{l}(n)=Softmax(logits^{l}(n))=logits(\Delta logits^{l}(n)+logits^{l-1}(n))
\end{equation}
Here,$logits^{l-1}(n)$ denotes the logits from preceding layer's distribution estimates, while $\Delta logits^{l}(n)$ represents the current layer's residual adjustment. The weighting function $W(n)$, implemented as differentiable look-up table, enables precision-adaptive offset scaling based on bin index $n$.

This architecture provides three key advantages: (1)Uncertainty-aware modeling: Explicit probability distributions capture positional ambiguity at multiple scales; (2)Progressive refinement: Residual logit adjustments enable gradual distribution sharpening across layers; (3)Geometry-aware adaptation: Bounding box dimensions $\left\{W_{c},H_{c}\right\}$ provide instance-specific scaling for offset magnitudes

Through this probabilistic formulation, CFDR effectively decouples coarse localization from fine boundary adjustment while maintaining differentiability throughout the refinement process. The hierarchical refinement mechanism allows the model to progressively resolve positional uncertainty, particularly benefiting ambiguous boundary regions where deterministic approaches typically fail.

\section{Experiments}
\label{sec:headings}
To validate the effectiveness of Contourformer, we conducted experiments on the SBD, COCO, and KINS datasets and discussed the results.
\subsection{Implementation Details}
Following the settings in D-FINE, we use HGNetv2\cite{lv2023detrs} B2 and B3 as the backbone networks, with the number of predicted vertices set to $N_{v}=64$ for object contour formation. We set the instance query size $N_{q}=300$, the number of sub-contours $N_{c}=8$, and the decoder layer numbers $l=4$ and $l=6$ (corresponding to the two different backbone networks, HGNetv2-B2 and HGNetv2-B3). All parameters in the deformable attention mechanism within the decoder layers follow the same settings as those in the D-FINE decoder. We adhere to the end-to-end paradigm of query-based instance detection, employing one-to-one sample matching. The bipartite matching is performed using the Hungarian algorithm, which considers both category predictions and contour point set similarity, with an additional optimization for bipartite matching based on the similarity between the contour's bounding box and the ground truth box. The model's encoder is responsible for generating initial detection boxes, thus only class prediction and bounding box cost are used for bipartite matching, and supervision is applied using classification loss and detection box loss. The loss for each decoder layer consists of three components: classification loss, point-to-point loss, and shape loss.
\begin{equation}
\mathcal{L}_{decoder}=\lambda_c\mathcal{L}_{cls}+\lambda_p\mathcal{L}_p+\lambda_s\mathcal{L}_{shape}
\end{equation}
$\lambda_c$, $\lambda_p$and $\lambda_p$ are weights used to balance different loss terms. The classification loss employs Variational Focal Loss (VFL)\cite{zhang2021varifocalnet} to achieve consistent constraints for both classification and localization of positive samples. The point-to-point loss supervises the location of each predicted point, implemented using L1 loss. The shape loss provides supervision at a higher edge-level geometry by computing the offsets $\nabla V=\left\{\Delta v^{1\rightarrow 0},v^{2\rightarrow 1},...,,v^{N_{v-2}\rightarrow N_{v-1}},v^{N_{v-1}\rightarrow 0} \right\}$ between adjacent points within each contour to define the shape of contour $V$. The loss is then calculated using cosine similarity between the ground truth offset $\nabla V_{gt}$ and the predicted offset $\nabla V_{pred}$ .
\begin{equation}
\mathcal{L}_{shape}=\sum_{i=0}^{N_{v}-1}cos\_sim(\Delta v_{pred}^{i+1\rightarrow i},\Delta v_{gt}^{i+1\rightarrow i})
\end{equation}
The weights $\lambda_c$, $\lambda_p$and $\lambda_p$ are set to 1.0, 1.0, and 0.25, respectively, to balance the different loss terms. We utilized the Adam optimizer for training all models and incorporated 100 denoising modules to accelerate the training process.

\subsection{SBD}

\begin{figure}[htbp]
    \centering
    \begin{subfigure}[t]{0.18\textwidth}
        \includegraphics[max height=4cm,keepaspectratio]{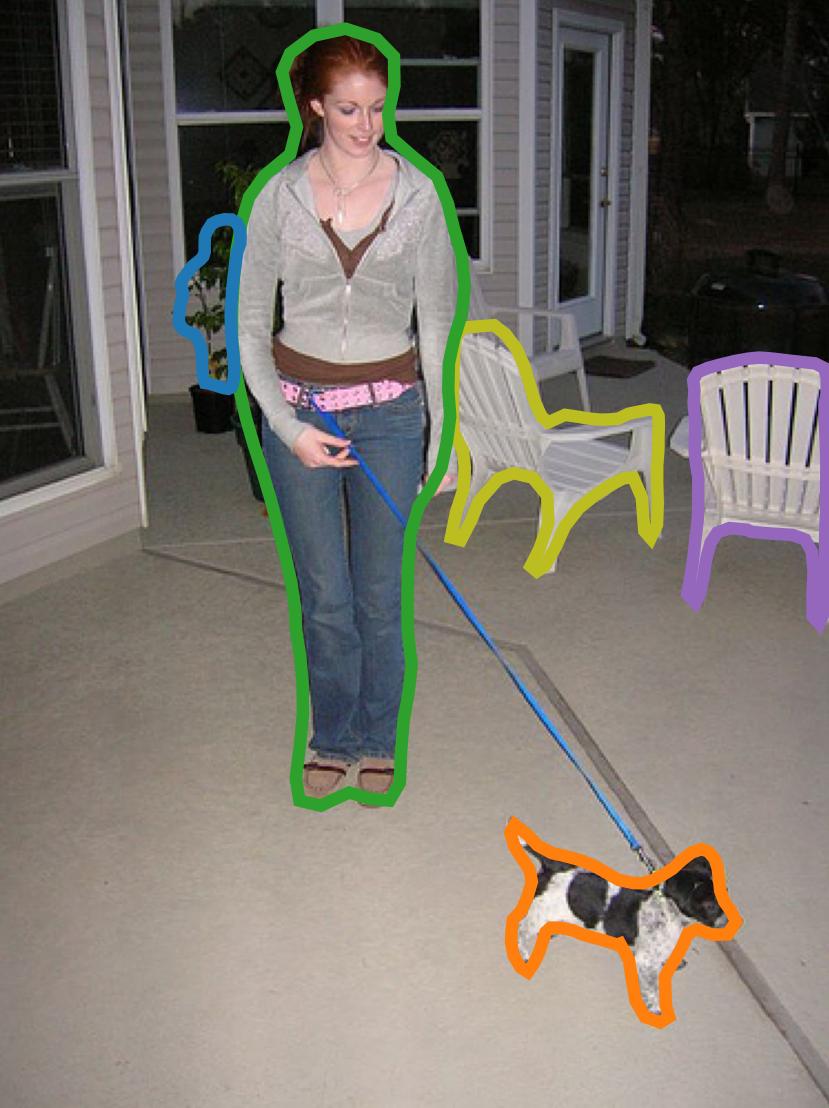}
        \label{fig:subfig1}
    \end{subfigure}
    \hfill
    \begin{subfigure}[t]{0.32\textwidth}
        \includegraphics[max height=4cm,keepaspectratio]{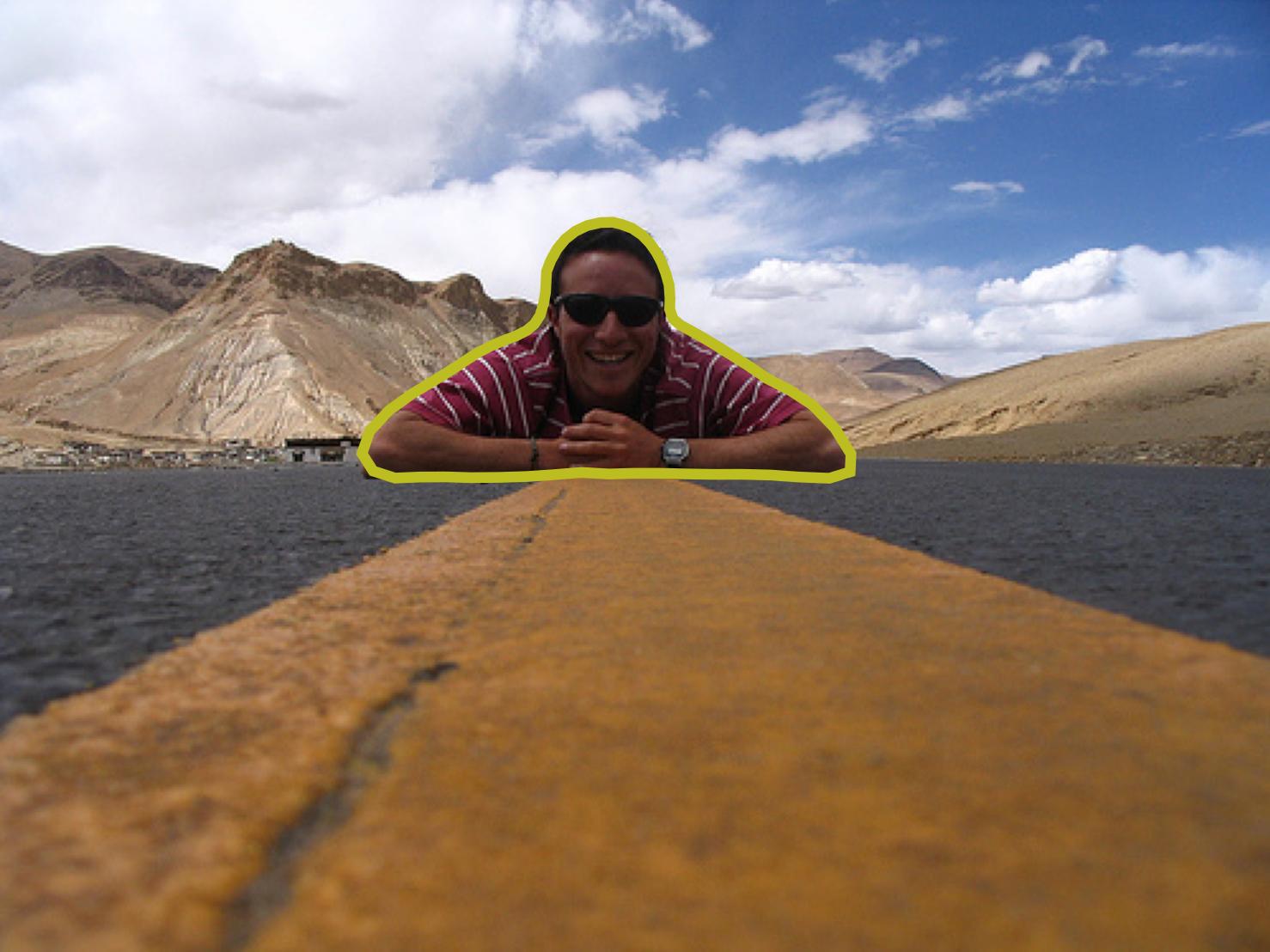}
        \label{fig:subfig2}
    \end{subfigure}
    \hfill
    \begin{subfigure}[t]{0.32\textwidth}
        \includegraphics[max height=4cm,keepaspectratio]{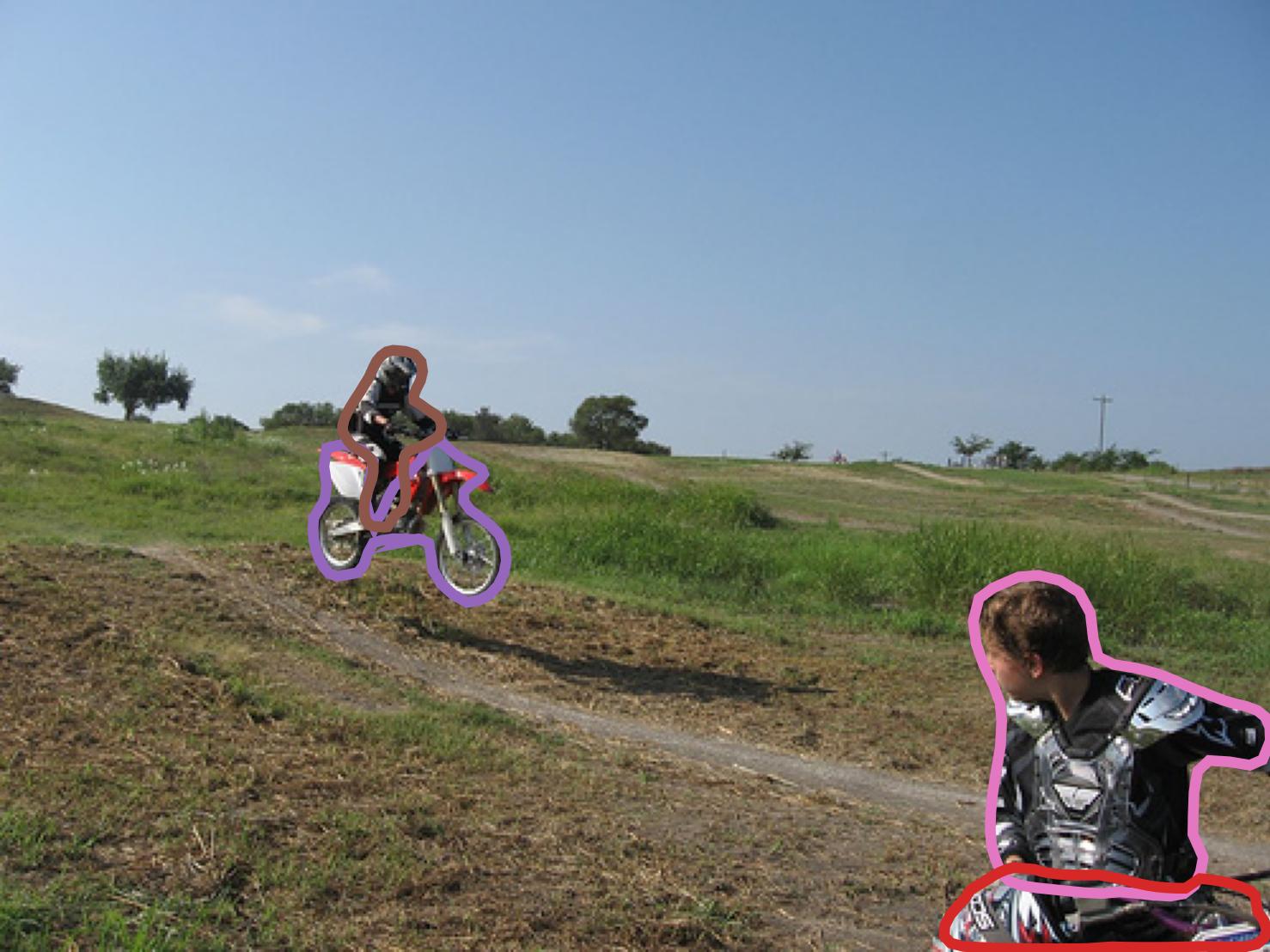}
        \label{fig:subfig3}
    \end{subfigure}
    \hfill
    \begin{subfigure}[t]{0.16\textwidth}
        \includegraphics[max height=4cm,keepaspectratio]{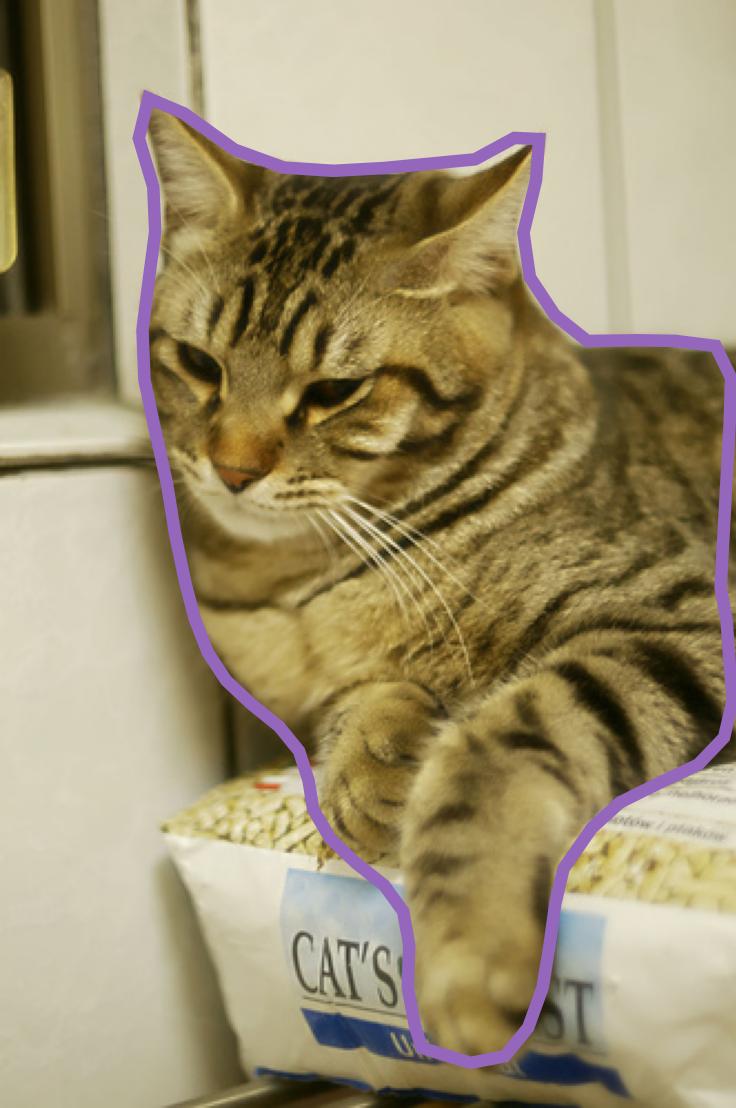}
        \label{fig:subfig4}
    \end{subfigure}

    \vspace{-0.3cm}

    \centering
    \begin{subfigure}[t]{0.155\textwidth}
        \includegraphics[max height=3.5cm,keepaspectratio]{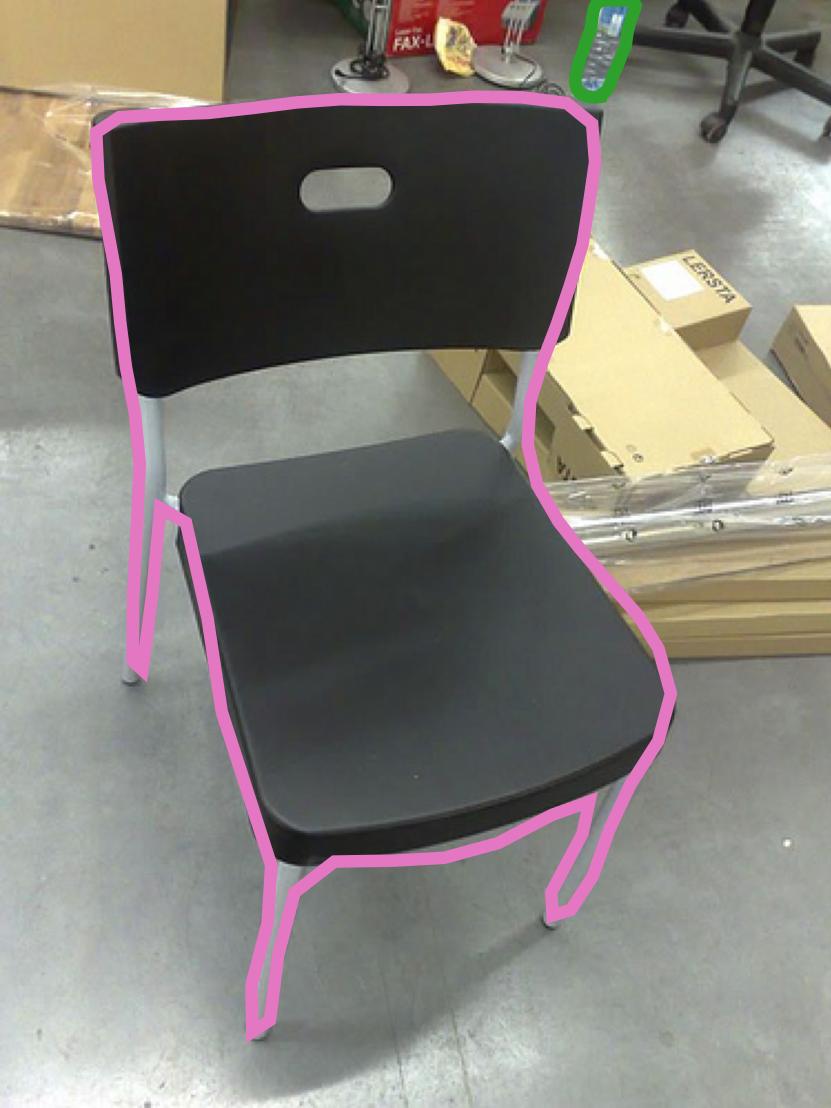}
        \label{fig:subfig5}
    \end{subfigure}
    \hfill
    \begin{subfigure}[t]{0.28\textwidth}
        \includegraphics[max height=3.5cm,keepaspectratio]{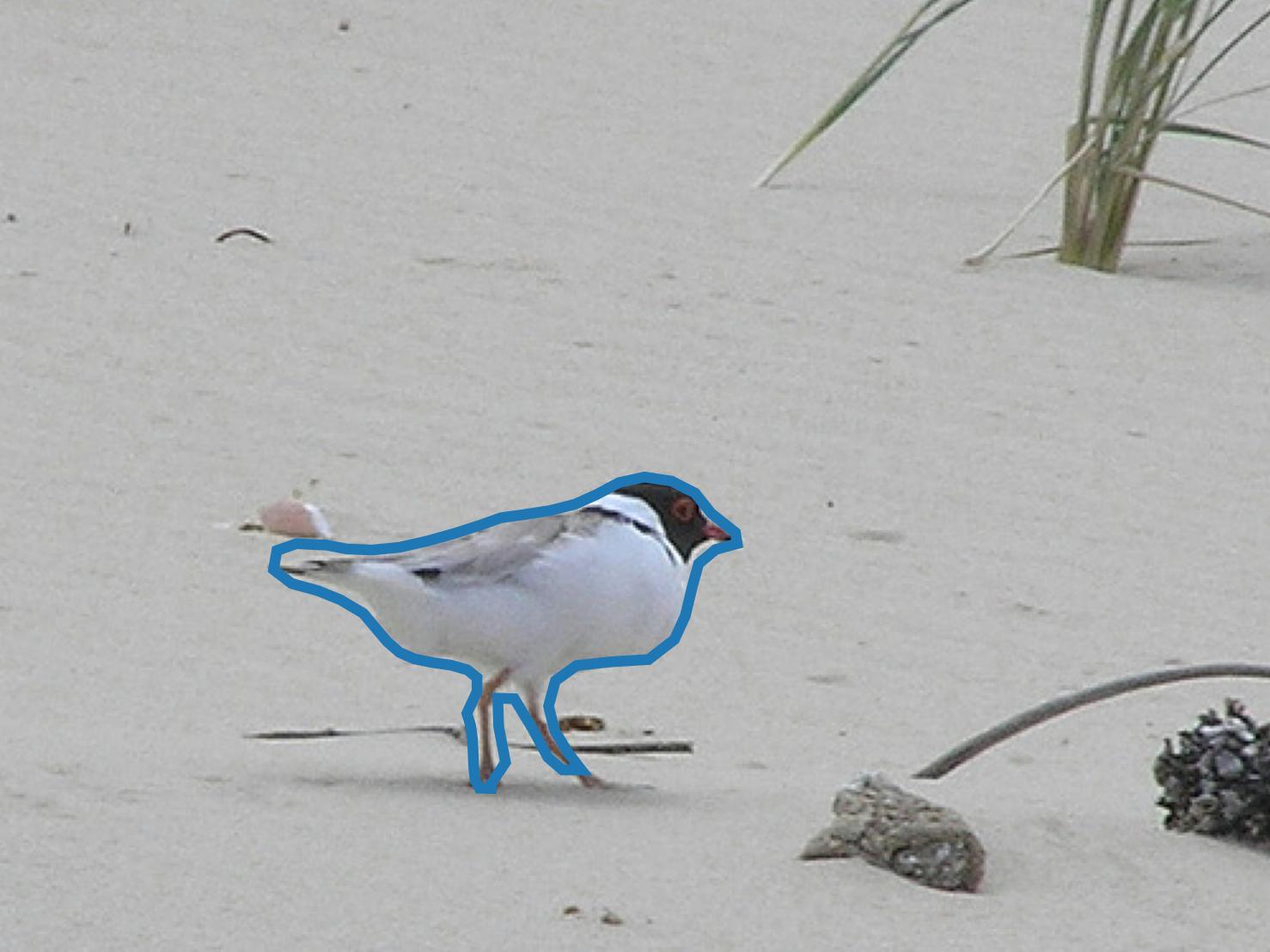}
        \label{fig:subfig6}
    \end{subfigure}
    \hfill
    \begin{subfigure}[t]{0.28\textwidth}
        \includegraphics[max height=3.5cm,keepaspectratio]{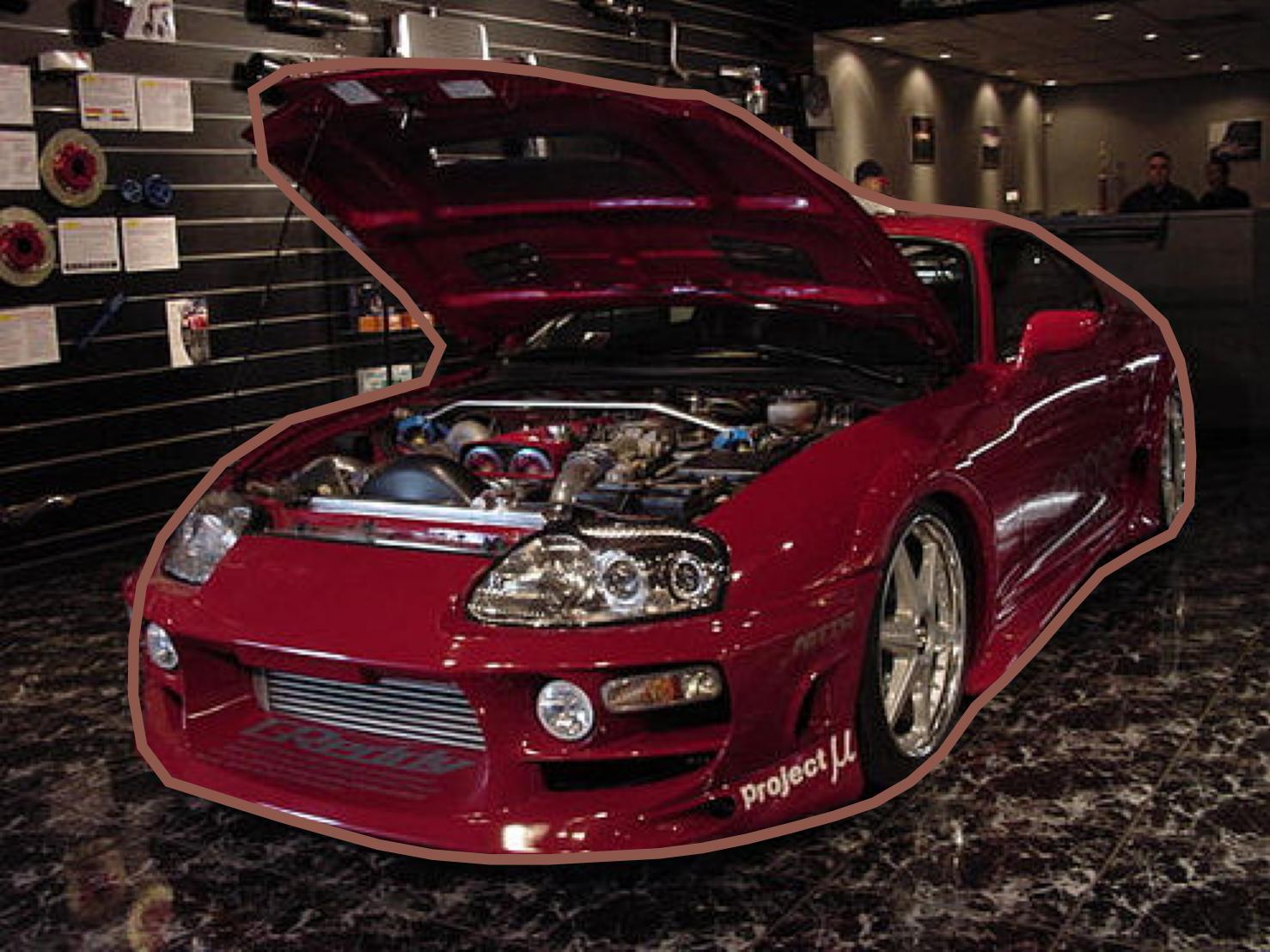}
        \label{fig:subfig7}
    \end{subfigure}
    \hfill
    \begin{subfigure}[t]{0.26\textwidth}
        \includegraphics[max height=3.5cm,keepaspectratio]{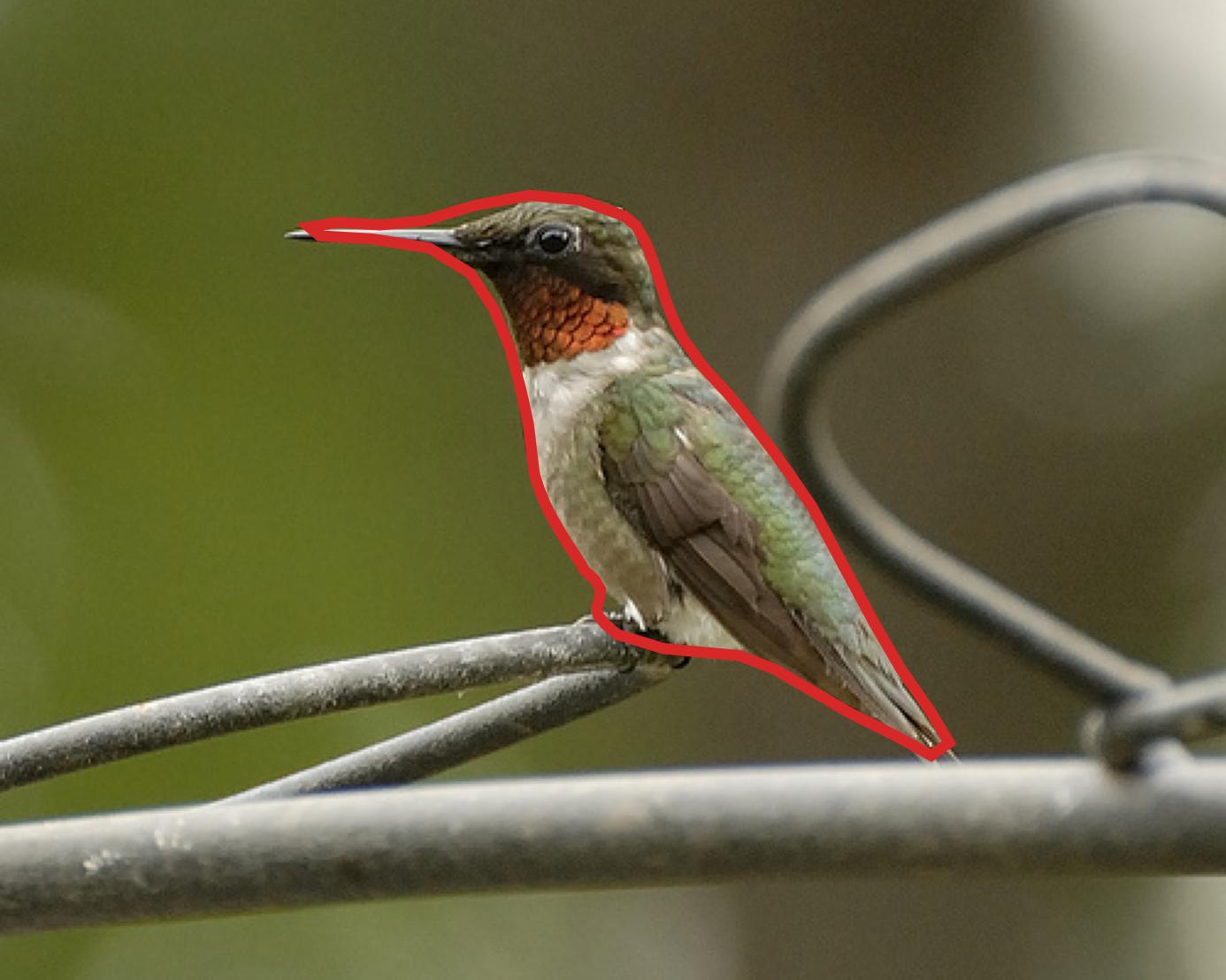}
        \label{fig:subfig8}
    \end{subfigure}
    \caption{Qualitative results of Contourformer on the SBD val set}
    \label{fig:sbd_val}
\end{figure}

The SBD dataset consists of 5,623 training images and 5,732 testing images, covering 20 distinct semantic categories. It leverages images from the PASCAL VOC\cite{2010The} dataset but re-annotates them with instance-level boundaries, providing a unique resource for evaluating contour detection and instance segmentation algorithms.
We report the performance of Contourformer and previous work based on the 2010 VOC $\mathrm{AP}_{vol}$, $\mathrm{AP}_{50}$, and $\mathrm{AP}_{70}$ metrics. $\mathrm{AP}_{vol}$ calculates the mean of average precision (AP) values across nine thresholds ranging from 0.1 to 0.9, offering a comprehensive evaluation of model accuracy over a range of IoU (Intersection over Union) thresholds.The network was trained and tested at a single scale of 512×512 following the preprocessing setup of polySnake, with training conducted for a total of 200 epochs. Some segmentation results are showcased in Figure \ref{fig:sbd_val}.

\begin{table}[htbp]
    \centering
    \caption{Performance comparison on the SBD val set: The '+' symbol denotes the utilization of HGNetv2 B3 as the backbone.}
    \label{tab:sbd_result}
    \begin{tabular}{lcccc}
        \toprule
        Method & Venue & $\mathrm{AP}_{vol}$ & $\mathrm{AP}_{50}$ & $\mathrm{AP}_{70}$ \\
        \midrule
        DeepSnake\cite{peng2020deep}   & $CVPR'2020$ & 54.4 & 62.1 & 48.3 \\
        DANCE\cite{liu2021dance}       & $WACV'2021$ & 56.2 & 63.6 & 50.4 \\
        EigenContours\cite{park2022eigencontours} & $CVPR'2022$ & - & 56.5 & - \\
        E2EC\cite{zhang2022e2ec}        & $CVPR'2022$ & 59.2 & 65.8 & 54.1 \\
        PolySnake\cite{feng2024recurrent}   & $TCSVT'2024$ & 60.0 & 66.8 & 55.3 \\
        TSnake\cite{hsutsnake}      & $APSIPA'2024$ & 60.5 & 67.2 & 56.1 \\
        Contourformer & - & 67.6 & 75.5 & 63.6 \\
        $\mathrm{Contourformer}^{+}$ & - & \textbf{69.2} & \textbf{76.9} & \textbf{65.8} \\
        \bottomrule
    \end{tabular}
\end{table}

In Table \ref{tab:sbd_result}, we compare the proposed Contourformer against other contour-based methods on the SBD val set. Our method achieves 69.2 $\mathrm{AP}_{vol}$, representing a significant improvement of 8.7 $\mathrm{AP}_{vol}$ over TSnake. This demonstrates that Contourformer not only outperforms existing methods in terms of accuracy but also maintains efficiency, capable of running at speeds of 24.6 FPS and 18.2 FPS at the 512×512 resolution during inference.

\subsection{COCO}

\begin{figure}[htbp]
    \centering
    \begin{subfigure}[t]{0.23\textwidth}
        \includegraphics[max height=2.8cm,keepaspectratio]{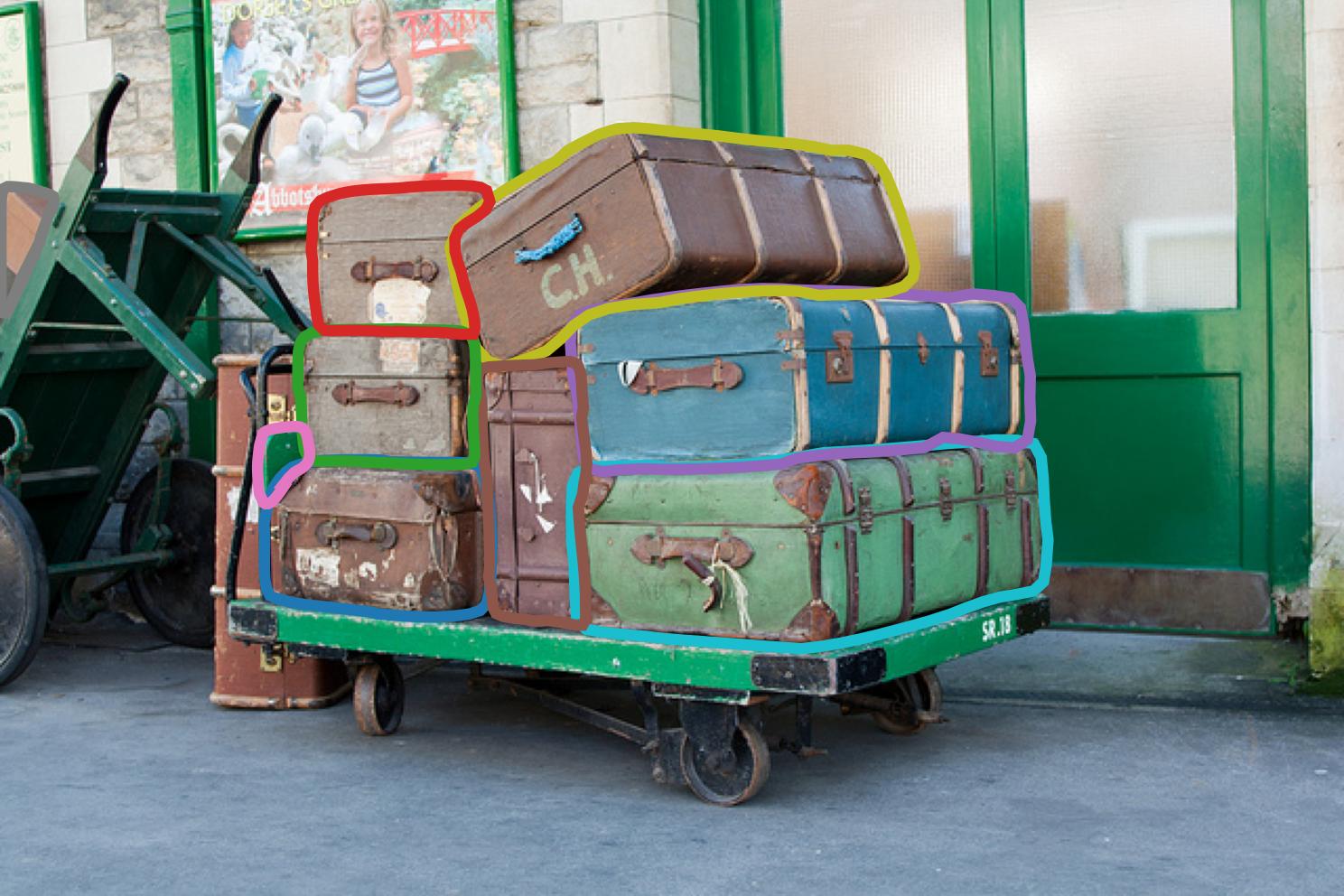}
        \label{fig:csubfig1}
    \end{subfigure}
    \hfill
    \begin{subfigure}[t]{0.23\textwidth}
        \includegraphics[max height=2.8cm,keepaspectratio]{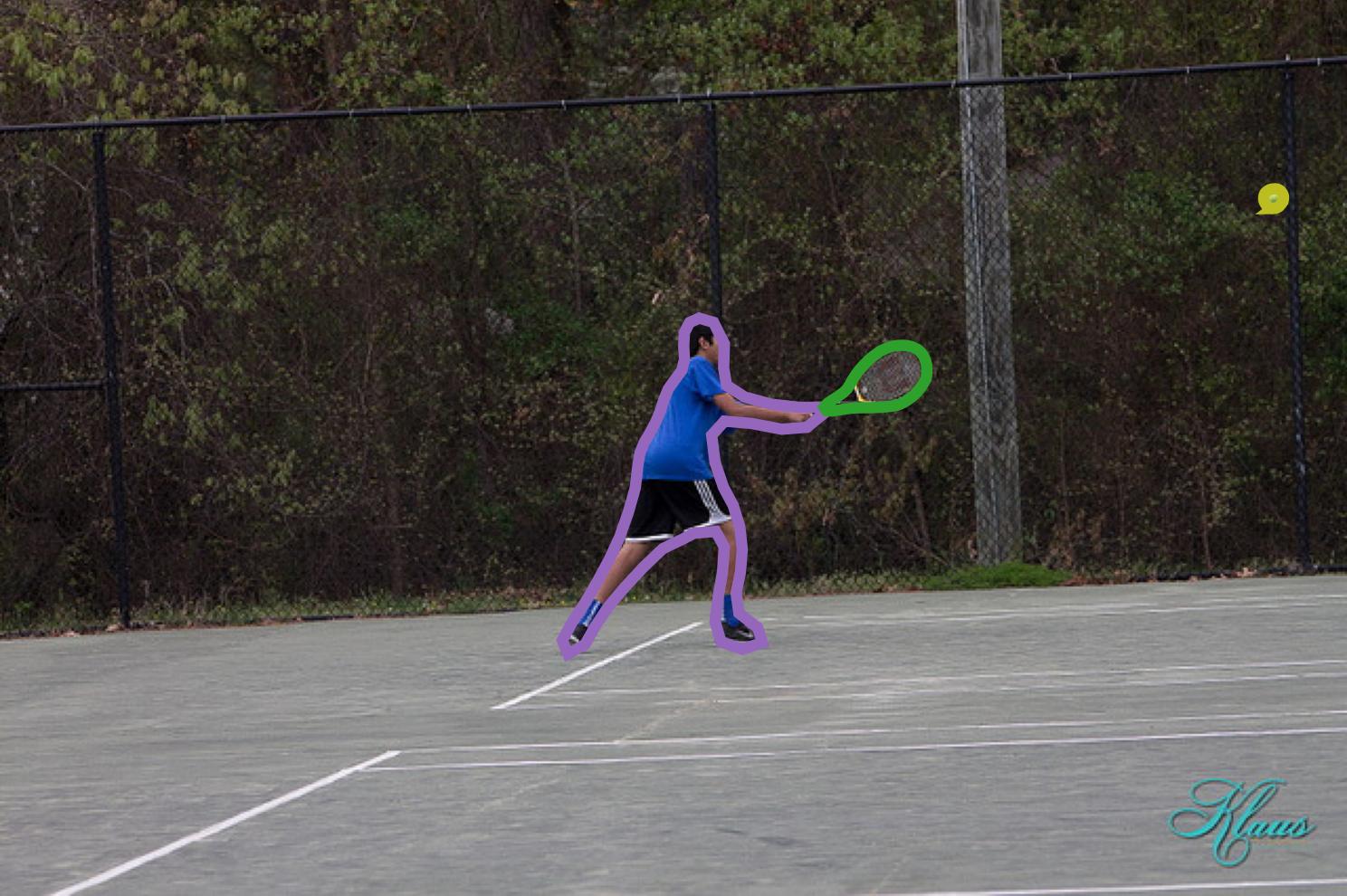}
        \label{fig:csubfig2}
    \end{subfigure}
    \hfill
    \begin{subfigure}[t]{0.23\textwidth}
        \includegraphics[max height=2.8cm,keepaspectratio]{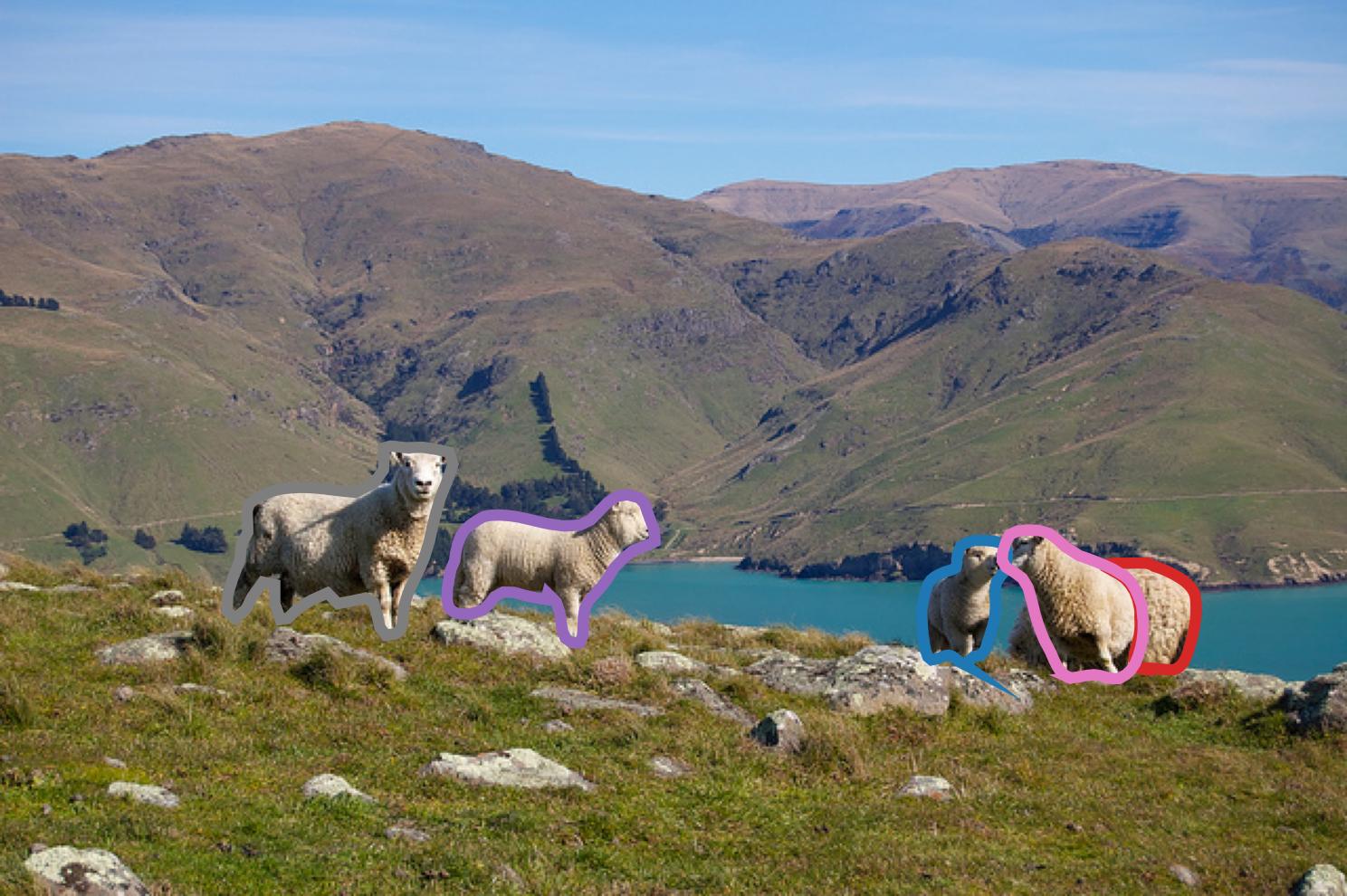}
        \label{fig:csubfig3}
    \end{subfigure}
    \hfill
    \begin{subfigure}[t]{0.23\textwidth}
        \includegraphics[max height=2.8cm,keepaspectratio]{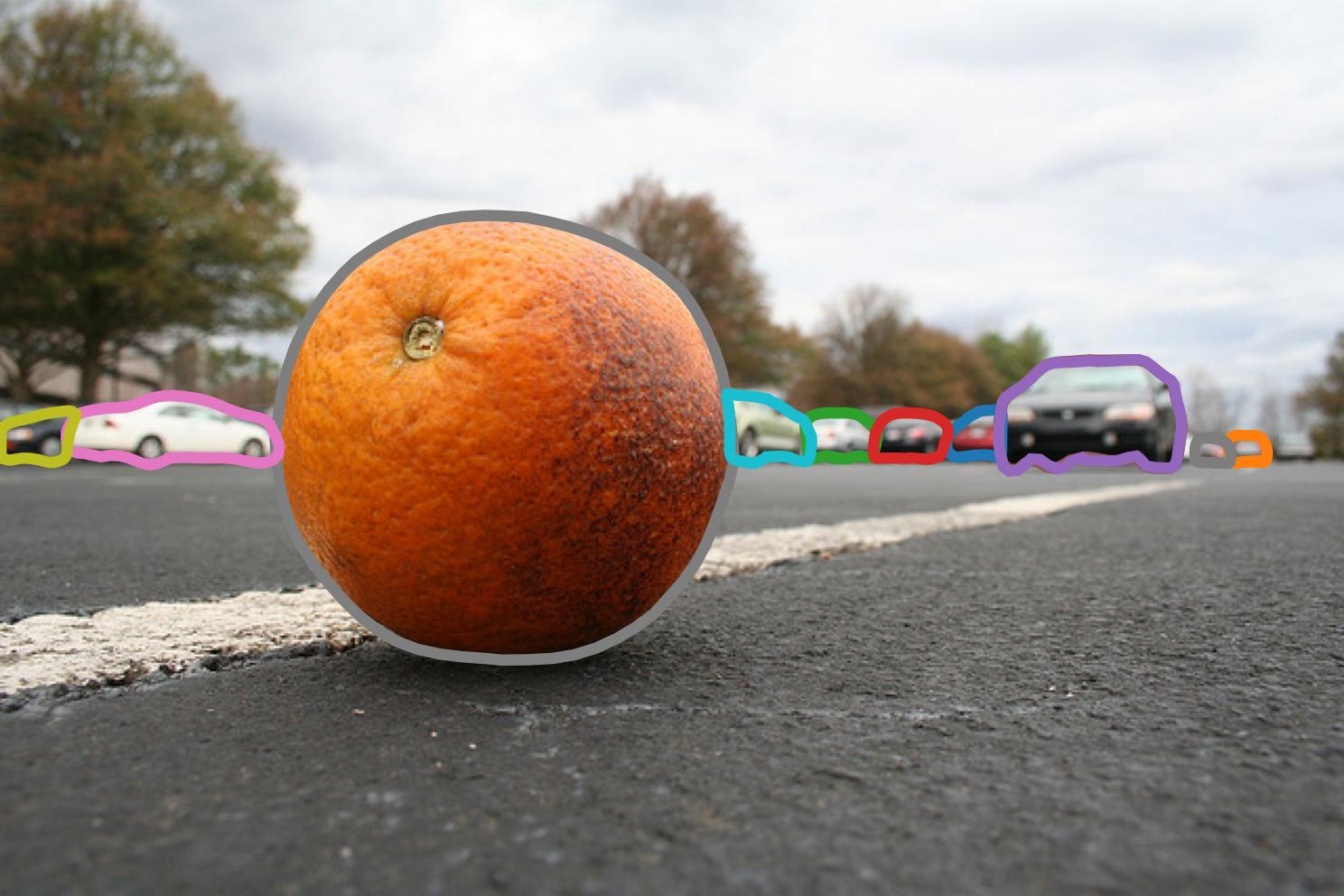}
        \label{fig:csubfig4}
    \end{subfigure}

    \vspace{-0.3cm}

    \centering
    \begin{subfigure}[t]{0.23\textwidth}
        \includegraphics[max height=2.8cm,keepaspectratio]{coco5.jpg}
        \label{fig:csubfig5}
    \end{subfigure}
    \hfill
    \begin{subfigure}[t]{0.23\textwidth}
        \includegraphics[max height=2.8cm,keepaspectratio]{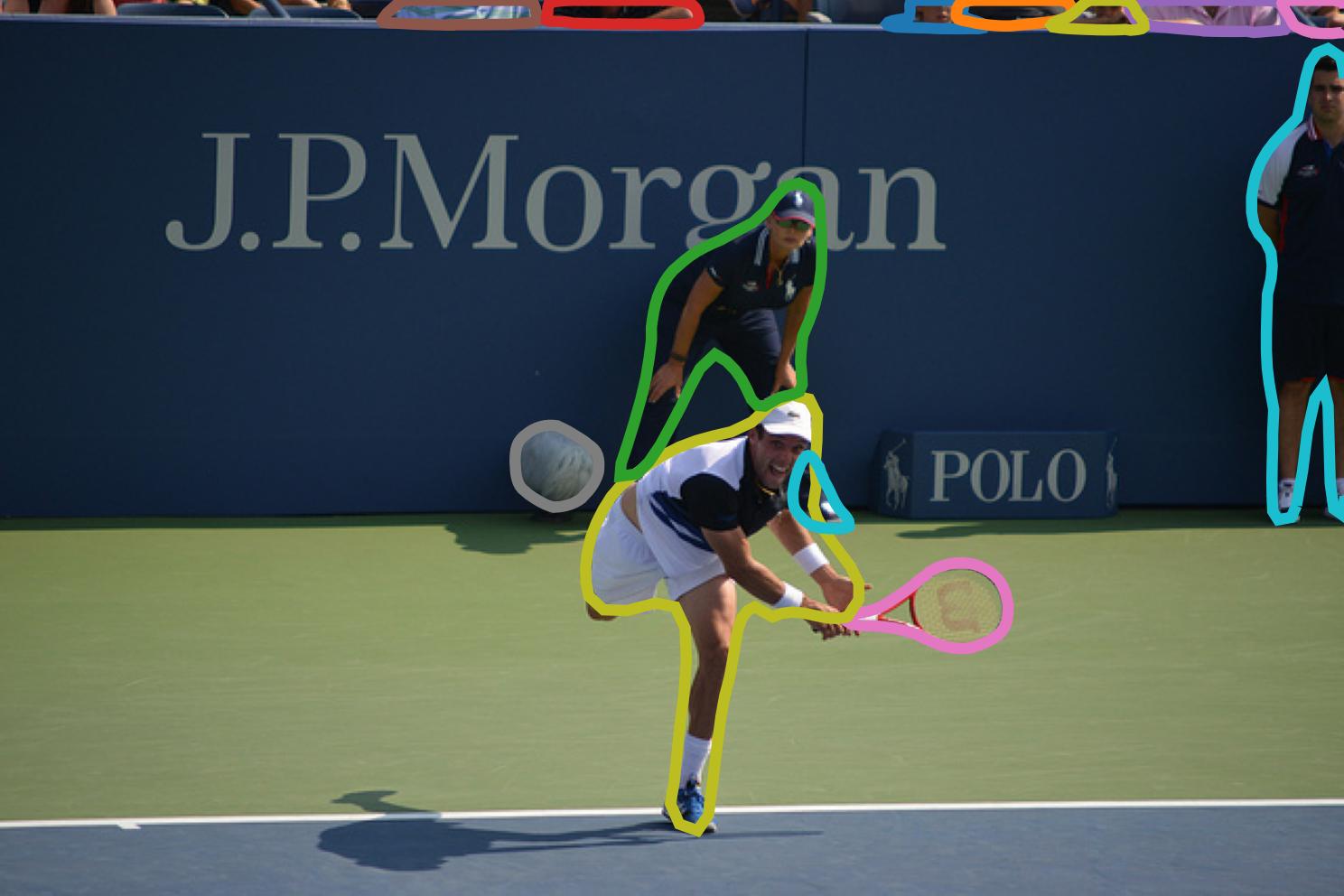}
        \label{fig:csubfig6}
    \end{subfigure}
    \hfill
    \begin{subfigure}[t]{0.23\textwidth}
        \includegraphics[max height=2.8cm,keepaspectratio]{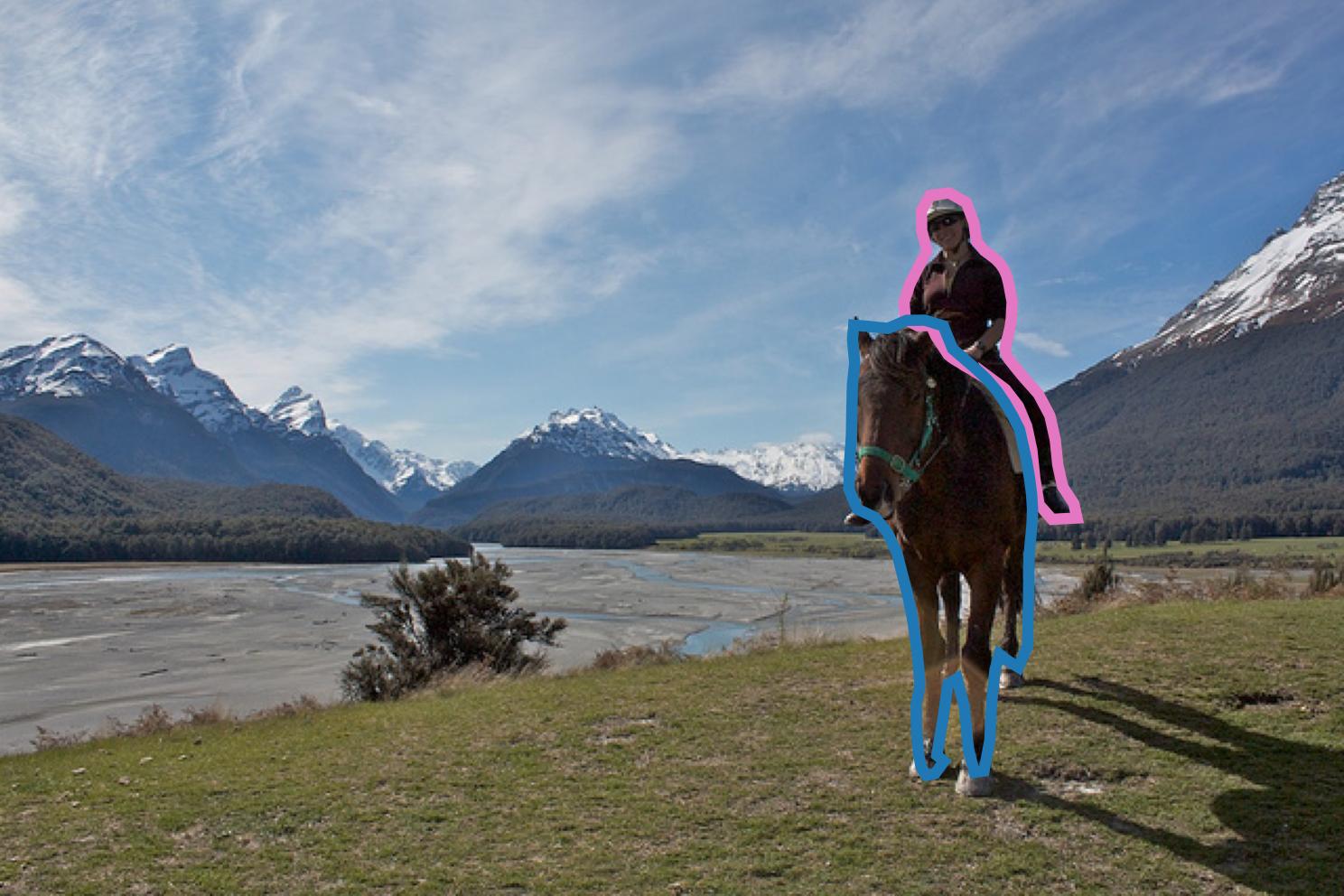}
        \label{fig:csubfig7}
    \end{subfigure}
    \hfill
    \begin{subfigure}[t]{0.23\textwidth}
        \includegraphics[max height=2.8cm,keepaspectratio]{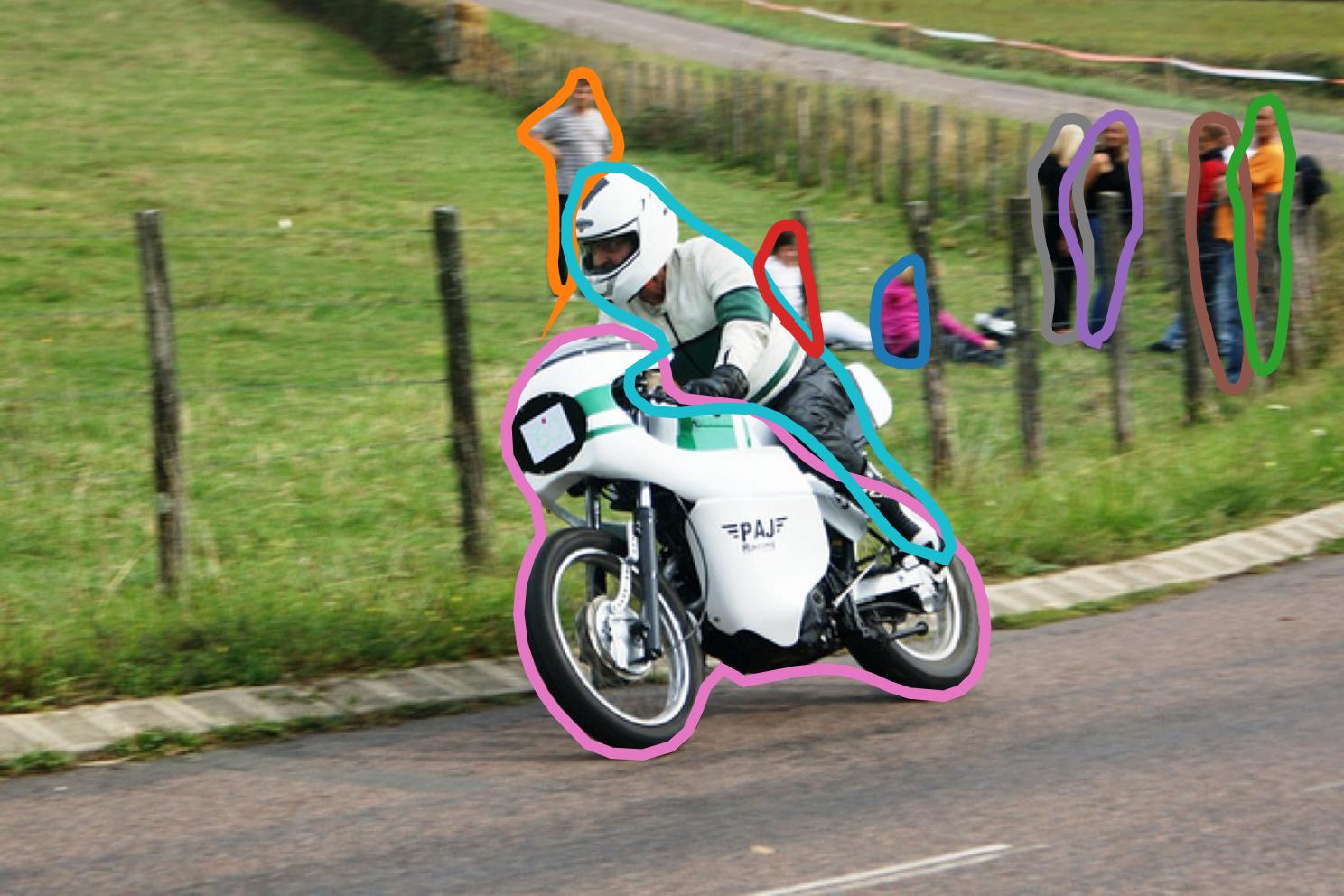}
        \label{fig:csubfig8}
    \end{subfigure}
    \caption{Qualitative results of Contourformer on the COCO val set}
    \label{fig:coco_val}
\end{figure}

The COCO dataset is a large-scale dataset containing 118,000 natural scene training images and 80 annotated foreground classes. Specifically, there are 115,000 images for training, 5,000 for validation, and 20,000 for testing. In our study, we adopt the COCO Average Precision (AP) metric as the evaluation standard. The network is trained and tested at a single resolution of 512×512, with a total training duration of 72 epochs. We present some segmentation results in Figure \ref{fig:coco_val}.

\begin{table}[http]
    \centering
    \caption{Performance comparison on the COCO val and test Set.}
    \label{tab:coco_result}
    \begin{tabular}{lcccc}
        \toprule
        Method & Venue & Backbone & $\mathrm{AP}_{val}$ & $\mathrm{AP}_{test-dev}$ \\
        \midrule
        DeepSnake\cite{peng2020deep}   & $CVPR'2020$ & DLA-34 & 30.5 & 30.3 \\
        PolarMask++\cite{9431650}       & $TPAMI'2021$ & R101-FPN & - & 33.8 \\
        E2EC\cite{zhang2022e2ec}        & $CVPR'2022$ & DLA-34 & 33.6 & 33.8 \\
        PolySnake\cite{feng2024recurrent}   & $TCSVT'2024$ & DLA-34 & 34.8 & 34.9 \\
        Contourformer & - & HGNetv2-B2 & 36.4 & 36.5 \\
        $\mathrm{Contourformer}^{+}$ & - & HGNetv2-B3 & \textbf{37.3} & \textbf{37.6} \\
        \bottomrule
    \end{tabular}
\end{table}

In Table \ref{tab:coco_result}, we compare the proposed Contourformer method with other contour-based approaches. Our method achieved 37.6AP in test-dev. Compared to PolySnake, the Contourformer improved performance by 2.7 AP.

\subsection{KINS}

\begin{figure}[htbp]
    \centering
    \begin{subfigure}[t]{0.3\textwidth}
        \includegraphics[max height=1.7cm,keepaspectratio]{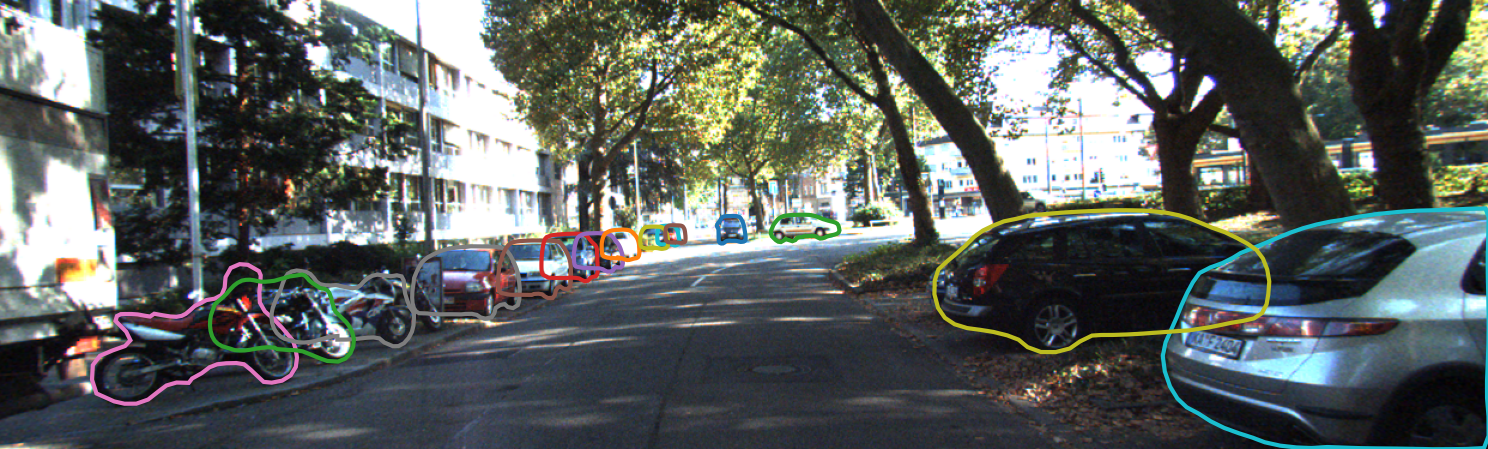}
        \label{fig:ksubfig1}
    \end{subfigure}
    \hfill
    \begin{subfigure}[t]{0.3\textwidth}
        \includegraphics[max height=1.7cm,keepaspectratio]{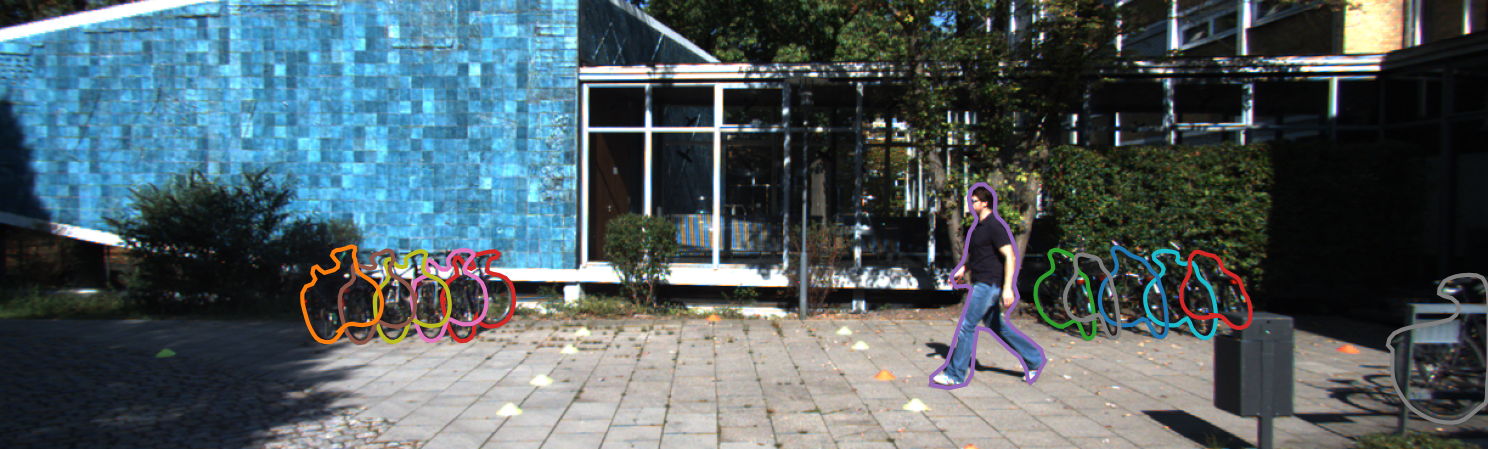}
        \label{fig:ksubfig2}
    \end{subfigure}
    \hfill
    \begin{subfigure}[t]{0.3\textwidth}
        \includegraphics[max height=1.7cm,keepaspectratio]{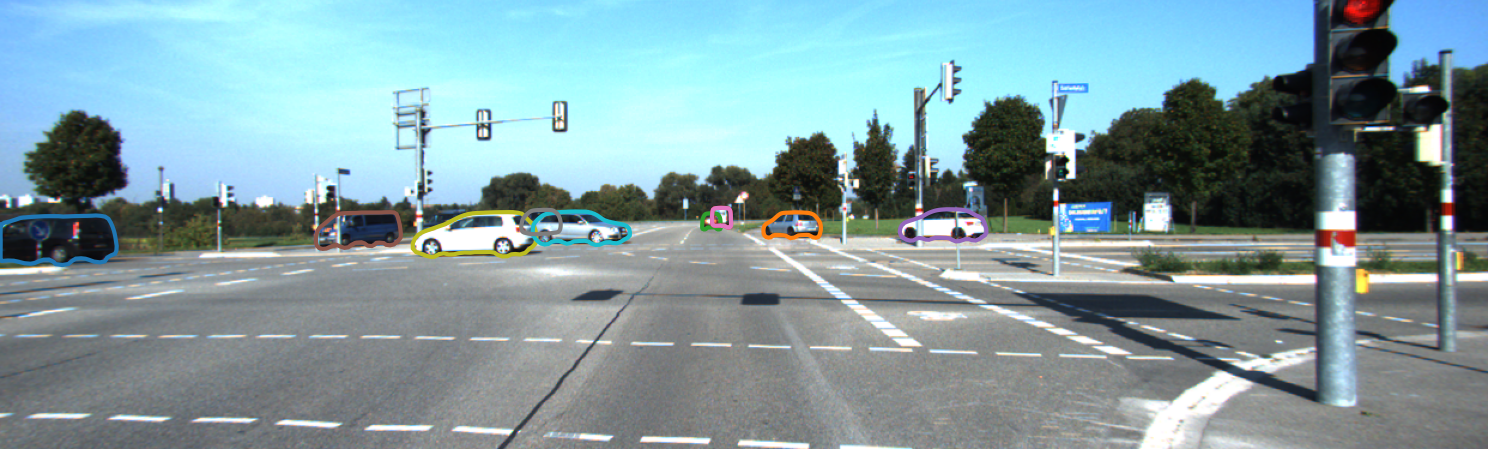}
        \label{fig:ksubfig3}
    \end{subfigure}

    \vspace{-0.3cm}

    \centering
    \begin{subfigure}[t]{0.3\textwidth}
        \includegraphics[max height=1.7cm,keepaspectratio]{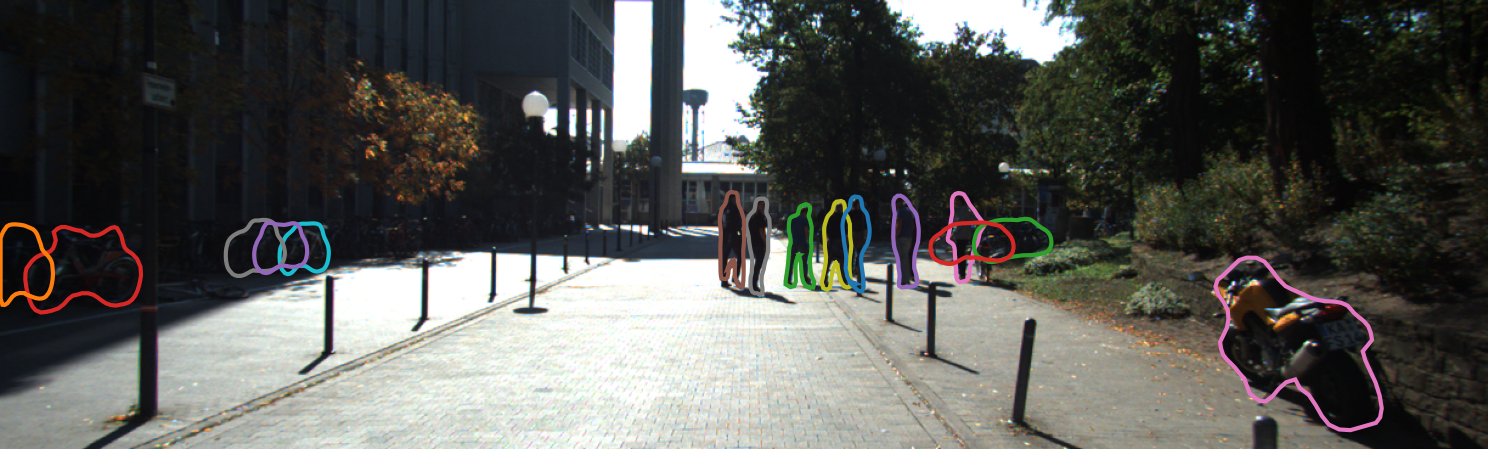}
        \label{fig:ksubfig5}
    \end{subfigure}
    \hfill
    \begin{subfigure}[t]{0.3\textwidth}
        \includegraphics[max height=1.7cm,keepaspectratio]{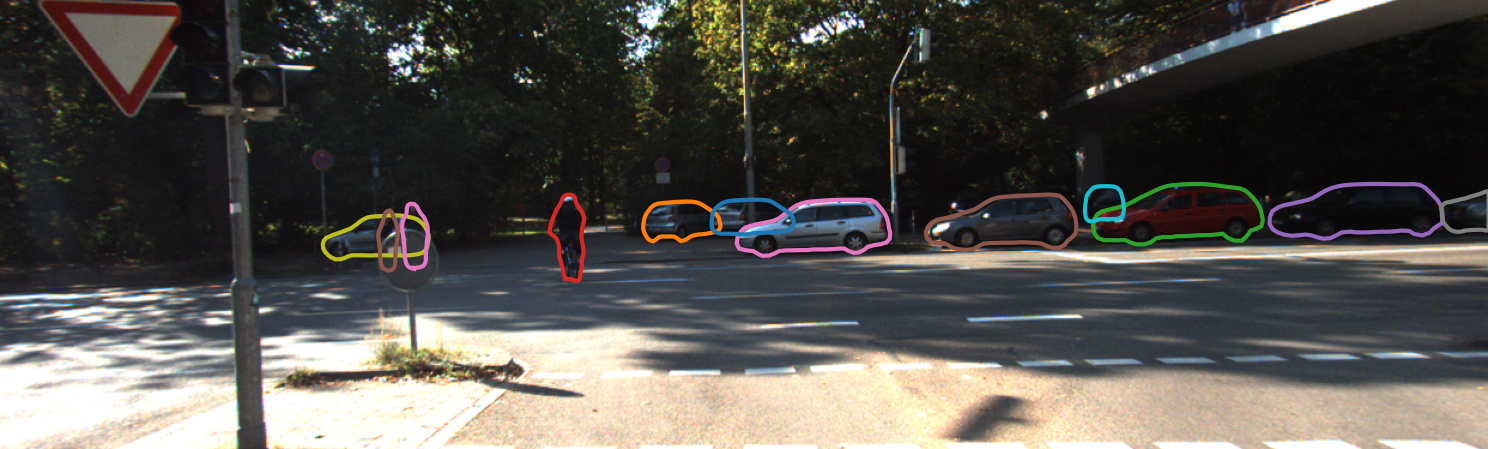}
        \label{fig:ksubfig6}
    \end{subfigure}
    \hfill
    \begin{subfigure}[t]{0.3\textwidth}
        \includegraphics[max height=1.7cm,keepaspectratio]{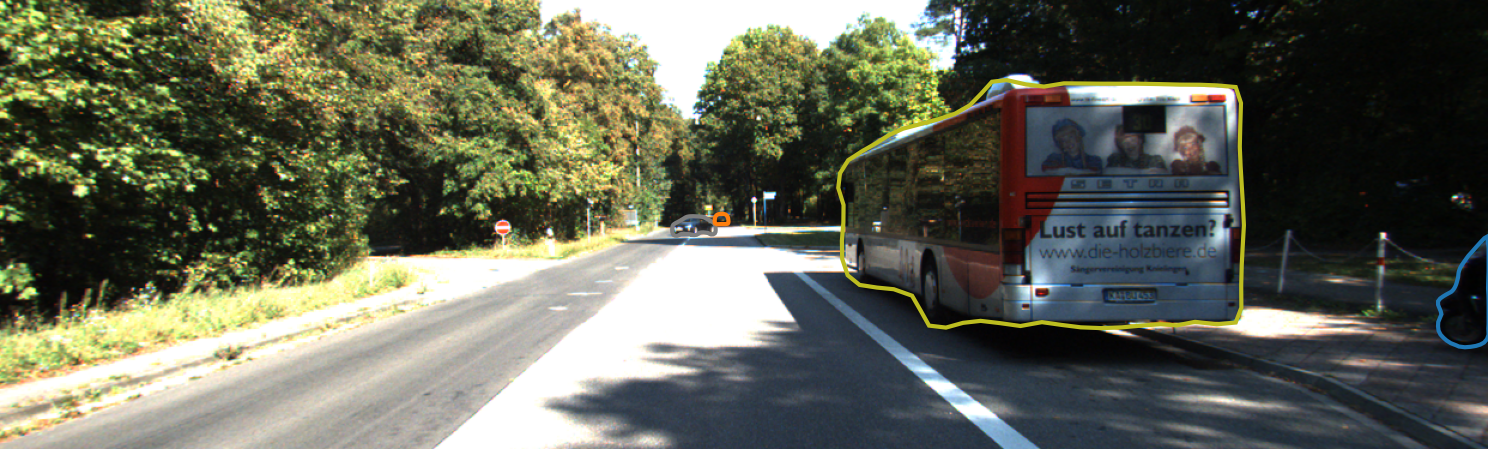}
        \label{fig:ksubfig7}
    \end{subfigure}

    \caption{Qualitative results of Contourformer on the KINS test set}
    \label{fig:kins_test}
\end{figure}

The KINS dataset uses images from the KITTI\cite{Geiger2013IJRR} dataset, annotated with instance-level semantic annotations. This dataset consists of 7,474 training images and 7,517 test images, divided into 7 foreground categories. We use the AP metric as the evaluation standard. The network was trained for 100 epochs and evaluated at a single resolution of 768×2496, following PolySnake's setup.

\begin{table}[htbp]
    \centering
    \caption{Performance Comparison on the KINS test Set.}
    \label{tab:kins_result}
    \begin{tabular}{lcc}
        \toprule
        Method & Venue & AP \\
        \midrule
        DeepSnake\cite{peng2020deep}   & $CVPR'2020$ & 31.3 \\
        E2EC\cite{zhang2022e2ec}        & $CVPR'2022$ & 34.0 \\
        PolySnake\cite{feng2024recurrent}   & $TCSVT'2024$ & 35.2 \\
        Contourformer & - & 35.3 \\
        $\mathrm{Contourformer}^{+}$ & - & \textbf{36.2} \\
        \bottomrule
    \end{tabular}
\end{table}

As shown in Table \ref{tab:kins_result}, we compare the proposed Contourformer method with other contour-based approaches. We present some segmentation results in Figure \ref{fig:kins_test}. The Contourformer runs at speeds of 14.4 FPS and 11.5 FPS at a resolution of 768×2496, achieving an improvement of 1.0 AP compared to PolySnake.

\subsection{Ablation Study}
To verify the effectiveness of the main components in our proposed Contourformer and their parameter selections, including the sub-contour decoupling mechanism, contour fine-grained distribution refinement, and the impact of the number of sub-contours ($N_{c}$) on model efficiency, we conducted a series of experiments. All ablation experiments were performed on the SBD dataset.

First, we validated the impact of the number of sub-contours ($N_{c}$) on model performance. As shown in Table \ref{tab:nc_number_select}, we tested four scenarios: $N_{c}$=0 (without using the sub-contour iteration optimization mechanism), 4, 8, and 16. The results indicated that as the number of sub-contours increased, the model's accuracy improved; however, this enhancement was accompanied by an increase in both inference time and memory usage. When $N_{c}=16$, the model achieved its highest accuracy, but this came at the cost of a significant rise in inference time and memory consumption. Conversely, when $N_{c}=8$, the model exhibited a relatively substantial improvement in accuracy without a considerable increase in inference time. Therefore, we selected $N_{c}=8$ as the optimal number of sub-contours for our model to balance accuracy and resource utilization effectively.

\begin{table}[http]
    \centering
    \caption{The Impact of the Number of Sub-Contours($N_{c}$) on Model Accuracy and Resource Consumption.}
    \label{tab:nc_number_select}
    \begin{tabular}{lccccc}
        \toprule
        Method & $\mathrm{AP}_{vol}$ & $\mathrm{AP}_{50}$ & $\mathrm{AP}_{70}$ & FPS & Memory Usage (MB) \\
        \midrule
        $N_{c}$=0 & 64.0 & 74.2 & 58.3 & 49.6 & 1632 \\
        $N_{c}$=4 & 65.1 \color{green}({+1.1}) & 74.3 \color{green}({+0.1}) & 60.3 \color{green}({+2.0}) & 39.8 \color{red}({-9.8}) & 2072 \color{red}({+440}) \\
        $N_{c}$=8 & 65.8 \color{green}({+0.7}) & 75.0 \color{green}({+0.7}) & 61.4 \color{green}({+1.1}) & 38.8 \color{red}({-1.0}) & 2640 \color{red}({+568}) \\
        $N_{c}$=16 & 66.5 \color{green}({+0.7}) & 75.6 \color{green}({+0.6}) & 62.1 \color{green}({+0.7}) & 28.3 \color{red}({-10.5}) & 3734 \color{red}({+1094}) \\
        \bottomrule
    \end{tabular}
\end{table}

Building on this foundation of $N_{c}=8$, we further investigated two additional aspects: whether employing sub-contour bounding boxes as deformable attention sampling ranges would influence model accuracy, and the effect of contour fine-grained distribution refinement on model precision. As demonstrated in Table \ref{tab:impact_sub_contours}, using sub-contour bounding boxes as deformable attention sampling ranges improved $\mathrm{AP}_{vol}$ by 1.3, without adversely affecting inference time or memory usage. Furthermore, incorporating contour fine-grained distribution refinement yielded an additional improvement of 0.5 in $\mathrm{AP}_{vol}$, albeit with a notable increase in inference time.These experiments provided valuable insights into the trade-offs between different components and parameter choices, guiding us toward an optimal configuration that maximizes performance while considering computational efficiency.

\begin{table}[htbp]
    \centering
    \caption{The Impact of Using Sub-Contour Bounding Boxes and Fine-Grained Contour Distribution Refinement on Model Accuracy and Resource Consumption}
    \label{tab:impact_sub_contours}
    \begin{tabular}{ccccccc}
        \toprule
        \multicolumn{2}{c}{Configuration} & $\mathrm{AP}_{vol}$ & $\mathrm{AP}_{50}$ & $\mathrm{AP}_{70}$ & FPS &Memory Usage (MB) \\
        \cmidrule(r){1-2} 
        Bounding Box & Refinement & & & & & \\
        \midrule
        & & 65.8 & 75.0 & 61.4 & 38.8 & 2640 \\
        \checkmark & & 67.1 \color{green}(+1.3) & 75.3 \color{green}(+0.3) & 63.3 \color{green}(+1.9) & 38.8\color{red}(-0) & 2640 \color{red}(+0) \\
        \checkmark & \checkmark & 67.6 \color{green}(+0.5) & 75.5 \color{green}(+0.2) & 63.6 \color{green}(+0.3) & 24.6\color{red}(-14.2) & 2716 \color{red}(+76) \\
        \bottomrule
    \end{tabular}
\end{table}

\section{Conclusion}

In this study, we propose Contourformer, a real-time contour-based instance segmentation algorithm. Our method is entirely grounded in the DETR paradigm and employs an iterative and progressive mechanism to incrementally refine contours, thereby enabling end-to-end inference capabilities. By introducing an innovative sub-contour decoupling mechanism and fine-grained distribution refinement techniques, Contourformer ensures stability and accuracy for each instance while maintaining segmentation speed.Experimental evaluations on standard datasets such as SBD, COCO, and KINS demonstrate that our approach significantly outperforms existing state-of-the-art methods, thus validating its effectiveness and versatility. We anticipate that Contourformer will serve as a pivotal baseline in this field, providing robust technical support for future research endeavors.


\end{document}